\documentclass[10pt,table]{article}
\usepackage[ieee]{amlpreprint}
\usepackage{fontspec}
\usepackage[bidi=default,english]{babel}
\babelprovide[import]{arabic}
\babelfont{rm}[BoldFont=texgyrepagella-bold.otf,ItalicFont=texgyrepagella-italic.otf,BoldItalicFont=texgyrepagella-bolditalic.otf]{texgyrepagella-regular.otf}
\babelfont{sf}[BoldFont=texgyreheros-bold.otf,ItalicFont=texgyreheros-italic.otf,BoldItalicFont=texgyreheros-bolditalic.otf]{texgyreheros-regular.otf}
\babelfont{tt}[BoldFont=lmmonolt10-bold.otf,ItalicFont=lmmono10-italic.otf,BoldItalicFont=lmmonolt10-boldoblique.otf]{lmmono10-regular.otf}
\babelfont[arabic]{rm}[BoldFont=Amiri-Bold.ttf,ItalicFont=Amiri-Italic.ttf,BoldItalicFont=Amiri-BoldItalic.ttf]{Amiri-Regular.ttf}
\usepackage{amsmath,amsfonts,booktabs,multirow,tabularx,graphicx,xcolor,microtype,url,enumitem}
\usepackage{xurl}
\usepackage{adjustbox}
\usepackage{longtable}
\usepackage[section]{placeins}
\usepackage{hyperref}
\usepackage{cleveref}
\makeatletter
\AtBeginDocument{\if@filesw
  \immediate\write\@auxout{\string\citation{paper-bstctl}}%
\fi}
\makeatother
\hypersetup{hidelinks}

\newcommand{\SuiteTaskCount}{43}
\newcommand{\SuiteFamilyCount}{9}
\newcommand{\SuiteItemCount}{133{,}689}

\newcommand{\AddedTaskSampleCap}{300}
\newcommand{\SharedRequestCount}{2{,}384}
\newcommand{\SharedMMMLUItemCount}{2{,}391}

\newcommand{\SystemCount}{67}
\newcommand{\DecisionSystemCount}{47}
\newcommand{\LLMSystemCount}{20}
\newcommand{\DeferredSystemCount}{4}

\newcommand{\RQOneTopMean}{71.81}
\newcommand{\RQOneJevMean}{69.24}
\newcommand{\RQOneOldJevMean}{68.77}
\newcommand{\RQOneTopJevGap}{2.57}

\newcommand{\CommonTaskCount}{40}

\newcommand{\ArabicJevMean}{69.18}
\newcommand{\ArabicOldJevMean}{68.71}
\newcommand{\ArabicTopMean}{71.75}
\newcommand{\ArabicMaxChangeBound}{0.332}
\newcommand{\RawOverlapChangeBound}{0.021}
\newcommand{\JevOverallRank}{6}

\newcommand{\RQOneJevKnowledge}{81.95}
\newcommand{\RQOneGemmaKnowledge}{80.19}
\newcommand{\RQOneJevNews}{70.57}
\newcommand{\RQOneGemmaNews}{68.65}
\newcommand{\RQOneLanguageGap}{10.01}

\newcommand{\RQTwoBelebeleGap}{-5.69}
\newcommand{\RQTwoSIBGap}{-0.35}
\newcommand{\RQTwoMMLUGap}{-3.63}
\newcommand{\RQTwoBankingGap}{-7.94}

\newcommand{\RQTwoExtensionDraws}{10{,}000}
\newcommand{\RQTwoExtensionPairs}{31{,}280}
\newcommand{\RQTwoExtensionItems}{40{,}618}
\newcommand{\RQTwoContrastCount}{69}
\newcommand{\RQTwoEgyptBelebeleGap}{-2.78}
\newcommand{\RQTwoMesoBelebeleGap}{-8.56}
\newcommand{\RQTwoSaudiMMLUGap}{-1.66}
\newcommand{\RQTwoMoroccoMMLUGap}{-7.24}
\newcommand{\RQTwoBankGemmaContrast}{-3.55}
\newcommand{\RQTwoBankJaisContrast}{-4.13}
\newcommand{\RQTwoBankJaisMSA}{20.37}
\newcommand{\RQTwoBankJaisDialect}{16.56}
\newcommand{\RQTwoMMLUOldMSA}{85.68}
\newcommand{\RQTwoMMLUNewMSA}{84.75}
\newcommand{\RQTwoMMLUOldDialect}{81.10}
\newcommand{\RQTwoMMLUNewDialect}{81.12}
\newcommand{\RQTwoSIBNonidenticalGap}{-0.39}
\newcommand{\RQTwoSIBNonidenticalPairs}{7{,}237}

\title{Evaluating Jev on Arabic: Comparing Decision Models and LLMs Across Arabic Tasks and Dialects}
\author[1,2,3]{Hossam Elsafty}
\author[2,3]{Tobias Deußer}
\author[2,3]{Lorenz Sparrenberg}
\author[2,3]{Rajkumar Ramamurthy}
\author[1,2,3]{Rafet Sifa}
\affil[1]{Fraunhofer Institute for Intelligent Analysis and Information Systems IAIS, Sankt Augustin, Germany}
\affil[2]{University of Bonn, Bonn, Germany}
\affil[3]{Lamarr Institute for Machine Learning and Artificial Intelligence, Germany}
\shorttitle{Evaluating Jev on Arabic}
\shortauthors{Elsafty et al.}
\date{}
\begin{document}
\maketitle
\begin{abstract}
How does Jev compare with other systems on Arabic tasks, and how does its performance change between Modern Standard Arabic (MSA) and dialects?
We evaluate \SystemCount{} systems, including \DecisionSystemCount{} decision models and \LLMSystemCount{} large language models (LLMs), on \SuiteTaskCount{} tasks across \SuiteFamilyCount{} task families using a shared zero-shot protocol.
The Arabic-focused frozen suite includes English controls in three parallel resources, which we also remove in a separate descriptive comparison.
Jev~1.14.0 achieves a mean task-primary metric of \RQOneJevMean{} on a scale multiplied by 100, compared with \RQOneOldJevMean{} for Jev~1.13.0 and \RQOneTopMean{} for Gemma-4-31B, the highest-scoring system on this aggregate.
Removing the recorded English controls, repeated cross-task items or both leaves the full system order unchanged.
Relative performance and gains between Jev versions vary across task families.
In content-matched comparisons within Jev~1.14.0, pooled dialect-minus-MSA accuracy gaps range from \RQTwoBankingGap{} to \RQTwoSIBGap{} percentage points across four datasets; dialect-specific results show larger variation, and SIB-200's source-clustered interval includes zero.
Paired contrasts with Jev~1.13.0, Gemma-4-31B and Jais-2-8B show a larger Palestinian banking penalty for Jev~1.14.0 than for either LLM under simultaneous uncertainty; the other gap contrasts remain inconclusive.
These descriptive findings show why aggregate scores should be read alongside task-family results, coverage and matched language comparisons.
They do not establish a general model ranking for dialect robustness or explain the causes of the observed gaps.
\end{abstract}

\section{Introduction}\label{sec:introduction}
Many language tasks require a decision with a fixed set of possible answers: assign an intent, identify a dialect, mark harmful content, or give a rating.
Jev addresses such tasks through typed questions rather than generated text.
An earlier study evaluated this interface across task types and compared Jev with large language models (LLMs) by scoring the probabilities of fixed answer codes \citep{deusser-etal-2026-jev-v2}.
This provides a starting point for comparing task decisions while allowing each system to retain its own inference interface.
The question for an Arabic-focused study is how well this comparison holds across the tasks and language varieties represented in its evaluation.

Arabic evaluation already includes broad language-understanding suites, knowledge questions and dialect resources.
ARLUE covers multi-dialectal language understanding, while ArabicMMLU evaluates subject knowledge through questions in Modern Standard Arabic (MSA) \citep{abdul-mageed-etal-2021-arbert,koto-etal-2024-arabicmmlu}.
DialectalArabicMMLU provides adapted versions of shared questions in Arabic dialects \citep{altakrori-etal-2026-dialectalarabicmmlu}.
These resources support different views of performance.
A broad task average summarizes the chosen suite, while a comparison of matched MSA and dialect questions examines changes for versions of the same source content.
Together, these views describe performance across the selected task inventory and changes across matched language varieties.

This paper asks two questions:
\begin{enumerate}[label=RQ\arabic*:]
\item How does Jev perform on Arabic tasks compared with other decision models and LLMs?
\item How does Jev's performance vary between MSA and Arabic dialects on content-matched tasks, and how do its accuracy gaps compare with those of selected LLMs?
\end{enumerate}
The first question covers overall performance, task-family differences, both Jev versions and coverage. The second describes paired accuracy differences by dataset and dialect, then compares Jev~1.14.0's gaps with Jev~1.13.0 and two LLMs selected by their overall RQ1 scores. We distinguish identifying a dialect from performing a task written in that dialect.

Both questions are descriptive. They were selected after evaluation from a broader preregistered study; the original plan and test records remain unchanged. \Cref{app:paper-scope} explains this scope and identifies the companion records for the original analyses.

The evaluation suite contains \SuiteTaskCount{} tasks in \SuiteFamilyCount{} families and compares \SystemCount{} systems: \DecisionSystemCount{} decision models and \LLMSystemCount{} LLMs.
The inventory counts evaluated versions or checkpoints separately, including both Jev versions.
Within this scope, the paper makes three contributions:
\begin{itemize}
    \item An organization of existing Arabic datasets into typed decision tasks, including parallel language varieties and a range of task label schemes.
    \item A shared zero-shot evaluation protocol that documents model interfaces, scoring, task support and answered-item coverage alongside performance.
    \item A comparison of models across Arabic task families and a source-aware analysis of matched MSA--dialect differences, including per-dialect results and paired contrasts with a fixed model panel.
\end{itemize}

Jev~1.14.0 has a mean task-primary score of \RQOneJevMean{} after multiplying by 100, above Jev~1.13.0's \RQOneOldJevMean{} and below Gemma-4-31B's \RQOneTopMean{}. The English-control removal sensitivity gives \ArabicJevMean{} for Jev~1.14.0 and \ArabicTopMean{} for Gemma-4-31B, with the full system order unchanged. Its relative performance changes across task families. The matched dialect comparison also varies by dataset, with a smaller pooled gap on SIB-200 than on the other retained sets. These results motivate reporting task profiles and matched language differences alongside the aggregate mean.

The comparison is bounded by the selected tasks, available answers and implemented readouts.
An average across different task metrics is not an overall accuracy, and a comparison based on different supported items needs separate coverage information.
The following sections place the study in prior work, describe the benchmark and protocol, and report each research question before discussing the findings and their limits.

\section{Related Work}\label{sec:related}
\subsection{Zero-Shot Classification and Decision Interfaces}\label{sec:related-decisions}
Zero-shot text classification can express label meanings in natural language.
Yin et al.~\citep{yin-etal-2019-benchmarking} studied this setting through a textual-entailment formulation across different label types.
For a generative large language model (LLM), a classification decision can instead be read from probabilities assigned to fixed answer codes.
The earlier Jev evaluation used this approach to compare open LLMs with Jev on shared requests \citep{deusser-etal-2026-jev-v2}.
Our open-LLM comparison follows this restricted-answer approach, with the supported codes and answered-item coverage made explicit in \cref{sec:setup}.

The earlier study also provides the methodological starting point for our use of Jev \citep{deusser-etal-2026-jev-v2}.
It evaluated a model that accepts a shared state and typed questions for option selection, ordered scoring and yes/no decisions, and examined both task performance and confidence.
We retain these question types while building an Arabic-focused comparison across decision models and LLMs.
This paper focuses on Arabic task performance and content-matched MSA--dialect differences. Returned probabilities provide the task decisions used by the shared scoring protocol.

\subsection{Arabic Evaluation and Dialect Coverage}\label{sec:related-arabic}
Arabic model evaluation already draws on suites that cover different tasks and language varieties.
ARLUE, introduced with ARBERT and MARBERT, combines datasets for multi-dialectal language-understanding evaluation and was used to compare fine-tuned models \citep{abdul-mageed-etal-2021-arbert}.
AlGhafa provides a multiple-choice benchmark for Arabic language models \citep{almazrouei-etal-2023-alghafa}.
These studies establish task coverage and evaluation format as distinct choices when comparing Arabic systems.
Our study uses a shared zero-shot protocol for Jev, other decision models and LLMs, with task-level results and coverage reported alongside aggregate scores.

Knowledge evaluation and dialect evaluation also address different questions.
ArabicMMLU uses school-examination questions in Modern Standard Arabic (MSA) across subject areas \citep{koto-etal-2024-arabicmmlu}.
DialectalArabicMMLU adapts MMLU-Redux questions into Arabic dialects, supporting comparisons across language varieties on matched question content \citep{altakrori-etal-2026-dialectalarabicmmlu}.
Dialect identification instead asks a model to recognize a text's variety, as in the country- and province-level tasks of NADI~2021 \citep{abdul-mageed-etal-2021-nadi}.
Success at identifying a dialect therefore answers a different evaluation question from success at answering questions written in that dialect.

We build on this evaluation tradition by representing the datasets in our suite as typed questions within one evaluation protocol.
The paper connects broad model and task-family comparisons to a focused within-system comparison of matched MSA and dialect content.
It also preserves the difference between parallel comparisons of shared content and analyses of independently collected native text, as detailed in \cref{sec:benchmark}.

\section{Benchmark Suite}\label{sec:benchmark}
\subsection{Scope and Selection}\label{sec:benchmark-scope}

We compare Jev with other decision models and large language models (LLMs) on existing Arabic classification and rating datasets.
The suite contains \SuiteTaskCount{} tasks in \SuiteFamilyCount{} families.
Parallel sets hold source content fixed across Modern Standard Arabic (MSA) and dialects, while other tasks evaluate decisions such as sentiment, intent and rating.
These controlled contrasts differ from evaluation on independently collected dialect text, where both content and language variety can change.

Tasks added during suite expansion followed a selection rule recorded before evaluation: declared licensing or usage terms, available reference labels, answers determined from the supplied input, and minimum sample-size and class-count requirements.
The rule excluded model-written text or labels unless model output was itself the object being evaluated.
These criteria apply to the additions, rather than uniformly to every original dataset.
The suite does not represent every Arabic task, dialect or application.

The frozen configurations of Belebele, SIB-200 and DialectalArabicMMLU include English controls alongside their Arabic varieties. These controls contribute to each resource's pooled primary score and therefore to the full-suite mean. The aggregate describes this frozen Arabic-focused inventory, rather than an Arabic-only subset. We therefore also report a descriptive sensitivity excluding the recorded English configurations; it retains Arabic and romanized-MSA examples under the same primary English-instruction template (T-EN) (\cref{app:language-overlap-sensitivity}).

\subsection{Task Families}\label{sec:task-families}

\Cref{tab:family-overview} summarizes coverage; \cref{app:task-inventory} lists every task, question type, primary metric and evaluation size.
Tasks within the suite measure different abilities: Belebele tests reading comprehension from a passage, whereas ArabicMMLU draws on school examinations across subjects \citep{bandarkar-etal-2024-belebele,koto-etal-2024-arabicmmlu}.
We therefore report task-level results alongside family summaries.
Family labels organize reporting; they are not an assumption of common difficulty or identical primary metrics.

\begin{table}[t]
\centering\small
\begin{tabularx}{\linewidth}{@{}llrX@{}}
\toprule
ID & Family & Tasks & Coverage \\
\midrule
A & Varieties and script & 4 & Reading and topic decisions across language varieties and scripts \\
B & Native dialect text & 3 & Dialect identification, sentiment, sarcasm, and stance \\
C & Label ladders & 5 & Tasks grouped around coarse-to-fine labels, including sentiment and intent \\
D & Application tasks & 12 & Intent, inference, relevance, similarity, moderation, and safety \\
E & Knowledge & 2 & Multiple-choice questions across subject areas \\
F & News and claims & 6 & Subjectivity, propaganda, claims, authorship, and truthfulness \\
G & Commonsense and culture & 4 & Cultural knowledge, plausibility, humor, and physical commonsense \\
H & Rating and grading & 4 & Readability and assessment of essays and short answers \\
I & Arabic language & 3 & Poetic meter, word meaning, and grammar \\
\midrule
 & Total & 43 & \\
\bottomrule
\end{tabularx}

\caption{Task families in the evaluation suite. Counts refer to tasks in our evaluation, not the number of labels, domains, or questions within each source dataset. Family membership organizes reporting; each research question uses its own specified tasks and conditions.}
\label{tab:family-overview}
\end{table}

\subsection{From Dataset Labels to Typed Questions}\label{sec:typed-questions}

Each item has an input, called the \emph{state}, and one or more named questions.
\textbf{Choice} selects an option from a specified set, \textbf{Noul} asks whether a property is present, and \textbf{Score} assigns a value on an ordered scale.
Choice and Score answers include probabilities over options or levels; Noul gives the probability of yes.
The evaluation protocol describes how each system supplies these answers.

\begin{figure}[t]
\centering
\includegraphics[width=\linewidth]{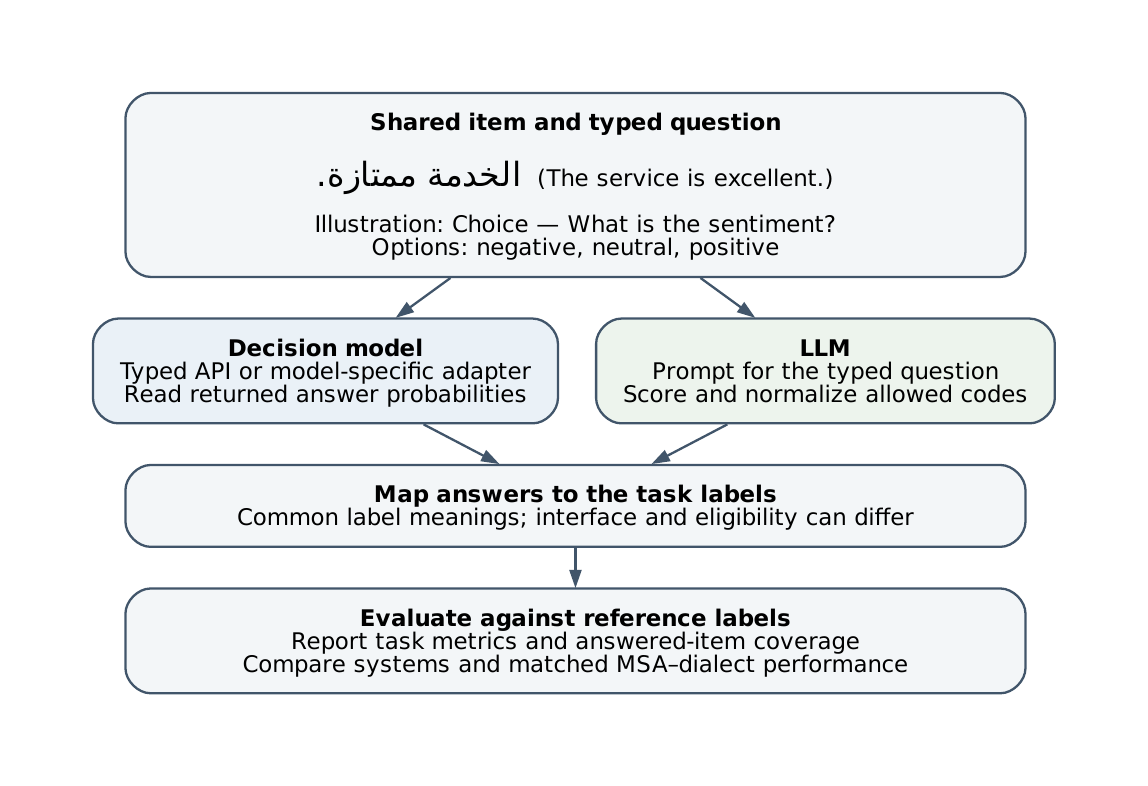}
\caption{Evaluation flow from a shared task representation through distinct model interfaces to label-based scoring. The Arabic sentiment example and English gloss are self-authored illustrations; no benchmark item, model prediction or measured probability is shown. Actual requests use frozen task wording, and supported questions and coverage depend on the system and condition.}
\label{fig:evaluation-flow}
\end{figure}

Several annotations can become separate questions about one state.
Our ArSarcasm-v2 mapping asks about sarcasm, sentiment and language variety for each tweet \citep{abu-farha-etal-2021-overview}.
Multilabel emotion classification uses one Noul question per emotion, allowing several emotions per item.
\Cref{fig:evaluation-flow} shows how the shared representation passes through the system interfaces to scoring.

Mapping choices define the label scheme being evaluated.
For BAREC readability, we use the seven-level scheme because its nineteen-level scheme exceeds Jev's supported Score range \citep{camel-lab-barec-sent}.
The evaluator uses numerical level indices, distinct from the human-readable level names.

\subsection{Evaluation Splits and Sampling}\label{sec:evaluation-items}

Each task uses the labelled evaluation split in its preregistered mapping.
For NADI 2021, this is the development split, with training data used for piloting \citep{abdul-mageed-etal-2021-nadi}.
Added tasks are capped at \AddedTaskSampleCap{} items, sampled within strata using a fixed hash order.
Their sampler deduplicates texts, excludes groups with conflicting labels and, where pilots are carved from the pool, removes those pilots from evaluation.
Original tasks retain their own split and sampling rules.

The frozen inventory contains \SuiteItemCount{} task-item entries across the suite.
These are not necessarily unique texts or independent observations, and the count does not establish that every system answered every question.
We report coverage with performance; the protocol and results state each comparison's pairing and resampling scope.

The primary task comparison uses each task's frozen evaluation inventory; paired dialect summaries retain only eligible matched records. Full task size, answered-item coverage and paired source count therefore have different meanings.

\subsection{Overlap and Label Constraints}\label{sec:data-constraints}

MMMLU Arabic \citep{openai-mmmlu} and DialectalArabicMMLU \citep{altakrori-etal-2026-dialectalarabicmmlu} share \SharedRequestCount{} distinct requests, affecting \SharedMMMLUItemCount{} MMMLU items because requests can recur within a dataset.
The registered task-level summaries retain this overlap; a separate descriptive sensitivity analysis removes the affected MMMLU items before recomputing overall means.
We report that analysis alongside the relevant comparisons.

Identical requests can also have conflicting reference labels, as in MADAR city identification \citep{bouamor-etal-2018-madar}.
A predictor assigning one fixed answer per identical request achieves its highest accuracy by choosing the most frequent label in each request group.
This bound helps interpret city accuracy; it does not directly bound other metrics or each realized run of a stochastic predictor.
These constraints inform the coverage, pairing and uncertainty rules described next.

\section{Systems and Evaluation Protocol}\label{sec:setup}
\subsection{Systems and Comparison Scope}\label{sec:systems-scope}

We evaluate \SystemCount{} systems: \DecisionSystemCount{} decision models and \LLMSystemCount{} LLMs, including pinned Jev versions \texttt{jev-1.13.0} and \texttt{jev-1.14.0}.
The comparison covers hosted decision services, Arabic and general-use open decision models, Arabic-centric LLMs, and the Qwen and Gemma families (\cref{tab:system-groups}).
\Cref{app:model-inventory} gives the systems and declared readouts; complete pinned revisions are supplied in the companion files.
Group differences are descriptive and cannot isolate effects of architecture or training.

\begin{table}[t]
\centering\small
\begin{tabularx}{\linewidth}{@{}lrX@{}}
\toprule
Group & Systems & Scope \\
\midrule
Hosted decision models & 4 & Jev versions and the Microsoft and OpenAI decision services \\
Arabic decision models & 5 & Arabic-focused releases in the pinned comparison \\
Other general decision models & 5 & Previously selected open decision-model releases \\
Arabic-centric LLMs & 9 & Jais, Jais 2, ALLaM and Fanar \\
Qwen LLMs & 6 & Qwen3.5 sizes and the Qwen3.8 comparison model \\
Gemma LLMs & 5 & Gemma 4 releases at the pinned sizes \\
Additional vendor decision models & 33 & Vendor releases added for the descriptive comparison \\
\midrule
Total & 67 & \\
\bottomrule
\end{tabularx}

\caption{Active systems grouped for presentation. Counts refer to evaluated versions or checkpoints, so related releases are counted separately. Arabic-centric and general-use labels follow the pinned study inventory. Group membership does not imply identical inference interfaces or coverage.}
\label{tab:system-groups}
\end{table}

\DeferredSystemCount{} pinned systems were deferred: \texttt{d3-turbo}, \texttt{h2o-lightning-4b}, \texttt{clef-omni}, and Jais-family-30B. They do not enter the reported inventory or its means. The companion records document these deferrals.

\subsection{Interfaces and Probability Readouts}\label{sec:readouts}

Each backend presents the shared state, questions and labels through its system's interface.
Hosted decision models receive typed service requests; open decision models use backend adapters that map returned probabilities to benchmark labels.
Choice selects the highest-probability option, Noul gives the probability of yes, and Score predicts the probability-weighted mean of ordered level indices.
Common scoring therefore allows different request serializations.

Open LLMs receive a prompt with the state, instructions and candidate answers.
The default readout scores single-token answer codes at the answer position, with thinking disabled in the chat template, then normalizes their probabilities over the allowed codes.
The distribution is conditional on those codes: it excludes other continuations and need not be calibrated to correctness.

Questions exceeding the context limit or unsupported by the default readout are recorded as skipped and affect coverage.

\subsection{Primary Condition and Eligibility}\label{sec:conditions}
The primary comparison uses the frozen T-EN task templates and the default readout described above. Frozen wording examples and the distinction between task templates and LLM wrappers appear in \cref{app:frozen-templates}. This fixed condition supports the reported model comparison; it does not establish performance under other prompt or answer representations.

\subsection{Metrics and Coverage}\label{sec:metrics-coverage}

Primary metrics include accuracy, macro-F1, positive-class F1, multilabel Jaccard similarity and rating Spearman correlation (\cref{tab:task-inventory}).
Accuracy measures the share of correct primary decisions. Positive-class F1 scores the task's named positive label; macro-F1 gives equal weight to the classwise F1 values for labels present in gold or predictions\citep{sklearn-f1-reference}. Multilabel tasks average the item-level Jaccard ratio of label-set intersection to union\citep{sklearn-jaccard-reference}; we assign one when both sets are empty and mark a label positive when its Noul probability is at least 0.5. Rating tasks use Spearman correlation between the predicted Score and source gold, measuring their rank association rather than exact rating agreement\citep{scipy-spearman-reference}.

Overall and family summaries are unweighted means of available task-primary scores, omitting missing or undefined scores.
These means combine different metrics and can cover different task sets across systems; per-task results and coverage are therefore needed to interpret them.

Task coverage is the number of available primary-metric item records divided by the frozen task size; mean coverage averages these fractions.
It measures primary-score support, not completion of every auxiliary question. For a multilabel primary metric, an item requires every binary primary head. The main results apply this complete-primary-item rule to the two refused AraTox heads found in one system, and retain the original export separately (\cref{app:primary-completeness-correction}).
We distinguish unsupported questions, missing responses and undefined metrics.
Paired comparisons further restrict eligibility to shared records.

\subsection{Statistical Protocol and Reproducibility}\label{sec:statistical-protocol}

The broader preregistration is available at \url{https://doi.org/10.5281/zenodo.23586552}. The paper's descriptive scope was chosen after evaluation and does not redefine its hypotheses or correction families (\cref{app:paper-scope}).

Both paper questions present descriptive comparisons selected after evaluation. Paper RQ1 reports overall and task-family scores, coverage and sensitivity summaries without confirmatory hypothesis-test claims. Paper RQ2 uses the same four matched datasets for both Jev versions, Gemma-4-31B and Jais-2-8B. The latter two have the highest overall available-task RQ1 means among general-use and Arabic-centric LLMs, respectively. Selection does not use dialect penalties, but remains conditional on the evaluated inventory and post-evaluation scope.

We join responses to exact frozen requests, then join systems by task and item identifier with matching source, variety, input and gold label. DialectalArabicMMLU source keys include the domain and question identifier. All four systems have valid primary answers for the same \RQTwoExtensionItems{} eligible item rows, producing \RQTwoExtensionPairs{} source--variety pairs. ArBanking77 first averages the same available paraphrases separately on each side of each source pair for every model; the matched analysis uses accuracy rather than its RQ1 macro-F1. No unsupported or missing answer is treated as an error in this panel.

Let $g_{m,d}=\operatorname{acc}_{m,d}-\operatorname{acc}_{m,\mathrm{MSA}}$ on the matched source population. A negative gap means lower dialect accuracy. We report each dialect separately and an equal-source--variety-pair average within each dataset. The cross-system contrast is $g_{\mathrm{Jev\,1.14},d}-g_{c,d}$ for comparator $c$; positive values mean a less negative gap for Jev. We do not pool different datasets. Whole articles are resampled for Belebele and SIB-200, and whole task/source identifiers for DialectalArabicMMLU and ArBanking77. Every draw keeps linked models, varieties and paraphrases together. The expanded analysis uses \RQTwoExtensionDraws{} draws per dataset and pointwise 95\% percentile intervals for accuracy gaps; fixed seeds and the earlier interval correction are documented in \cref{app:rq2-extension-method,app:dialect-interval-correction}.

For cross-system contrasts we also report approximate simultaneous 95\% bands covering the complete \RQTwoContrastCount{}-contrast grid, including dialect-specific and pooled rows. Within each dataset, bands use the 98.75th percentile of the maximum absolute centered bootstrap contrast divided by its bootstrap standard deviation; the radius is that critical value times the contrast's standard deviation. Bonferroni allocation across four datasets accounts for their shared content without assuming independence. These new descriptive bands do not replace the original preregistered RQ2 tests or correction family. A separate exploratory sensitivity removes pairs with identical full state/question inputs across MSA and dialect versions; its intervals are pointwise.

All current primary T-EN scores, item counts and coverage fractions were reproduced from the existing responses (\cref{app:full-primary-validation}). This verifies the primary numerical reporting, rather than historical server settings or secondary confidence analyses. Original test outputs and qualifications remain in the companion archive (\cref{app:paper-scope}).

The system ranking is descriptive; these means alone do not establish statistical separation.

Frozen task mappings, templates and item identifiers specify intended inputs; the inventory records pinned model revisions.
Generated tables and source hashes trace the manuscript's statements to those records and saved outputs.
\Cref{app:runtime-provenance} defines the recorded per-request timing and processing rates reported beside the evaluation results.
The companion files retain full model-source metadata, audit evidence and original test records.

\section{Results}\label{sec:results}
\subsection{Paper RQ1: Overall and Task-Family Performance}\label{sec:rq1}
Gemma-4-31B has the highest available-task mean at \RQOneTopMean{} on the scale multiplied by 100. Jev~1.14.0 scores \RQOneJevMean{}, above Jev~1.13.0's \RQOneOldJevMean{} and \RQOneTopJevGap{} points below Gemma. Jev~1.14.0 ranks \JevOverallRank{}th among all \SystemCount{} systems. \Cref{tab:rq1-summary} shows both Jev versions, the other hosted decision services and the highest-mean system from each original presentation group, with recorded processing rates beside their scores; the full ranking is in \cref{tab:all-system-results-timing}. Means use available primary scores, so task and item coverage qualify this ordering.

\begin{table}[!htbp]
\centering\small
\begin{tabularx}{\linewidth}{@{}Xrrrrr@{}}
\toprule
System & Score & Records/s & Arabic-only & Tasks & Coverage (\%) \\
\midrule
Gemma-4-31B & 71.81 & 7.86 & 71.75 & 43/43 & 100.00 \\
h2o-lightning-31b & 71.12 & 7.11 & 71.07 & 43/43 & 100.00 \\
jev-1.14.0 & 69.24 & 4.93 & 69.18 & 43/43 & 100.00 \\
OpenAI-Decisions-gpt-6-luna & 68.79 & 3.57 & 68.72 & 43/43 & 99.84 \\
jev-1.13.0 & 68.77 & 5.22 & 68.71 & 43/43 & 100.00 \\
SII-Decision-27B & 68.39 & 2.06 & 68.31 & 43/43 & 100.00 \\
Microsoft-Decision-1 & 66.06 & 2.79 & 65.97 & 43/43 & 100.00 \\
Qwen3.5-27B & 65.96 & 6.42 & 65.88 & 43/43 & 100.00 \\
Qwen3.8-27B & 65.15 & 6.53 & 65.05 & 43/43 & 100.00 \\
ara-jev-Faisal & 59.93 & 3.65 & 59.76 & 40/43 & 93.02 \\
Jais-2-8B & 54.54 & 36.68 & 54.46 & 43/43 & 100.00 \\
Jais-family-13B & 46.84 & 1.37 & 46.75 & 43/43 & 99.60 \\
\bottomrule
\end{tabularx}

\caption{Fixed-group summary ordered by unrounded score. Scores are mean task-primary metrics multiplied by 100. Records/s uses recorded processing time under each backend's execution settings (\cref{app:runtime-provenance}). Arabic-only removes English controls; tasks and coverage describe the frozen evaluation population.}
\label{tab:rq1-summary}
\end{table}

A descriptive sensitivity comparison after the support correction uses the same \CommonTaskCount{} tasks with finite scores for every system in the fixed summary group (\cref{app:support-sensitivity}). Two close comparisons change order: Jev~1.13.0 versus OpenAI Decisions, and Qwen3.5-27B versus Microsoft Decision-1. This task-level restriction does not ensure identical answered items. Removing the recorded English controls gives means of \ArabicJevMean{} for Jev~1.14.0 and \ArabicOldJevMean{} for Jev~1.13.0, compared with \ArabicTopMean{} for Gemma-4-31B. Removing repeated MMMLU items changes any system's overall mean by less than \RawOverlapChangeBound{} aggregate points. The complete 67-system order is unchanged under English-control removal, overlap removal or both; these descriptive checks do not establish statistical separation (\cref{app:language-overlap-sensitivity}).

\begin{figure}[t]
\centering
\includegraphics[width=\linewidth]{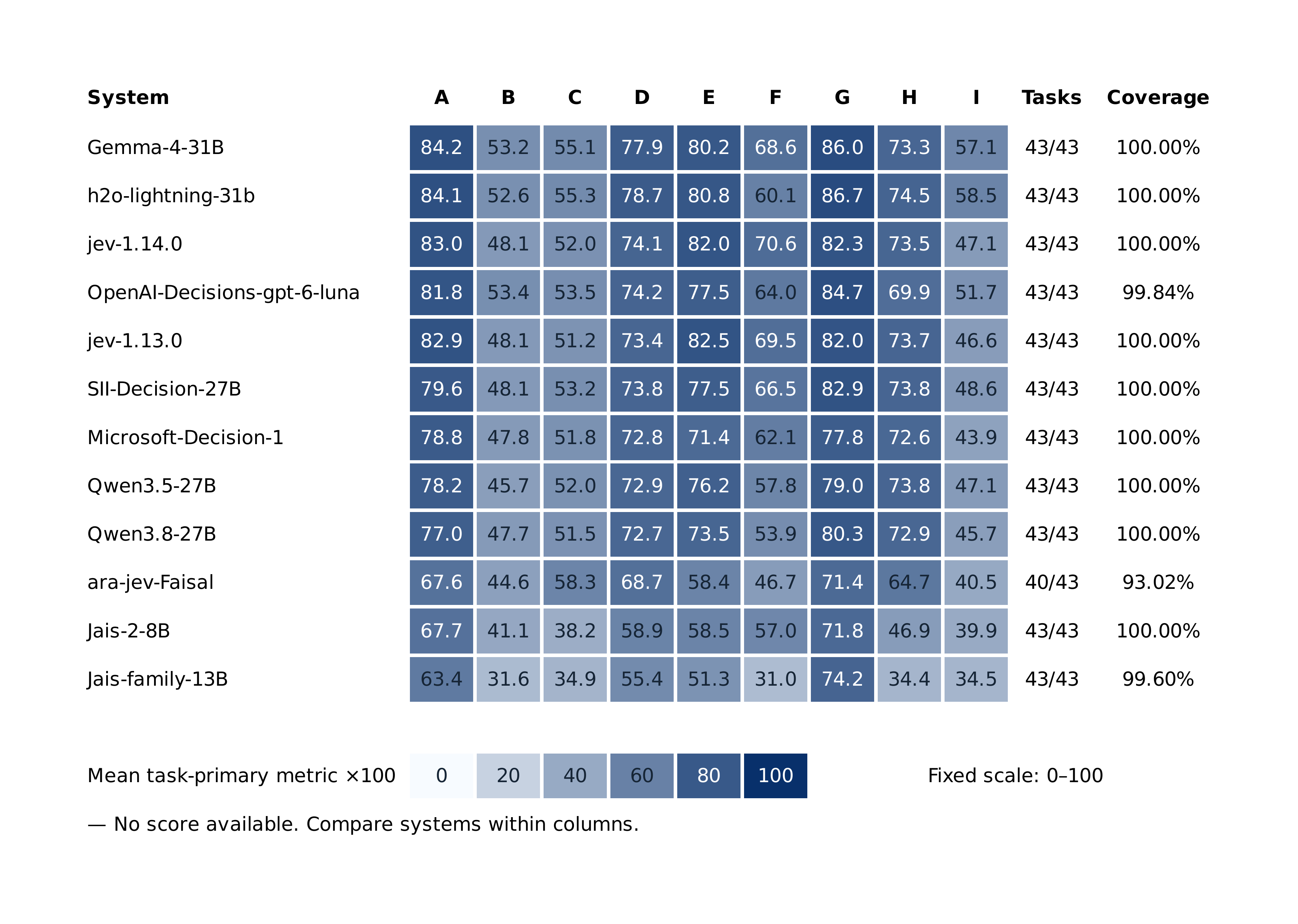}
\caption{Task-family means multiplied by 100 for the fixed summary population, ordered by overall mean. Families: A, varieties/script (including English controls); B, native dialect; C, label ladders; D, applications; E, knowledge; F, news/claims; G, commonsense/culture; H, rating/grading; I, Arabic language. Colors use a fixed 0--100 scale. Family means use available task scores; tasks and coverage summarize the evaluation population. Numerical validation is described in \cref{app:full-primary-validation}.}
\label{fig:family-heatmap}
\end{figure}

\Cref{fig:family-heatmap} shows that the aggregate ordering varies across task families, using the same selected systems as \cref{tab:rq1-summary}.
The detailed comparison of both Jev versions with Gemma-4-31B and h2o-lightning-31b is retained in \cref{tab:rq1-families}; the two comparators have the highest overall saved means among the LLMs and non-Jev decision models, respectively.
Jev~1.14.0 exceeds Gemma-4-31B in the saved knowledge mean (\RQOneJevKnowledge{} versus \RQOneGemmaKnowledge{}) and news-and-claims mean (\RQOneJevNews{} versus \RQOneGemmaNews{}).
Its largest family-level deficit against Gemma is in Arabic language tasks, at \RQOneLanguageGap{} aggregate points.
These contrasts describe variation in relative performance within families; they do not establish that one family is intrinsically harder than another.

The change between Jev versions is also uneven across families.
The saved Jev~1.14.0 mean is lower than Jev~1.13.0 for knowledge and rating/grading, while the other family means are higher.
The overall increase therefore does not represent a gain in every family.

These comparisons describe overall and task-family performance rather than testing the original hypotheses. The companion archive preserves the original RQ1 test records and their unresolved qualifications (\cref{app:paper-scope}).

\FloatBarrier
\subsection{Paper RQ2: Content-Matched MSA and Dialect Performance}\label{sec:rq2}
\paragraph{Matched accuracy across datasets.}
Jev~1.14.0 has lower pooled dialect accuracy than MSA accuracy on Belebele, DialectalArabicMMLU and the corpus-linked Palestinian subset of ArBanking77. The respective gaps are \RQTwoBelebeleGap{}, \RQTwoMMLUGap{} and \RQTwoBankingGap{} percentage points, with pointwise source-aware intervals below zero. SIB-200's pooled gap is \RQTwoSIBGap{} points and its interval includes zero (\cref{tab:rq2-paired-gaps}). This small average does not establish equivalence in every dialect. The table also shows absolute accuracy for the fixed model panel, so a small gap can be distinguished from strong performance on both versions.

\begin{table}[!htbp]
\centering\small
\begin{tabularx}{\linewidth}{@{}Xrrr@{}}
\toprule
System & MSA (\%) & Dialect (\%) & Gap [95\% interval] \\
\midrule
\multicolumn{4}{@{}l}{\textbf{Belebele}: 4{,}500 pairs; 488 clusters} \\
Jev 1.13.0 & 95.44 & 89.33 & -6.11 [-7.24, -4.97] \\
Jev 1.14.0 & 95.56 & 89.87 & -5.69 [-6.81, -4.57] \\
Gemma-4-31B & 94.44 & 89.56 & -4.89 [-6.13, -3.65] \\
Jais-2-8B & 82.11 & 73.20 & -8.91 [-10.90, -6.95] \\
\addlinespace
\multicolumn{4}{@{}l}{\textbf{SIB-200}: 8{,}032 pairs; 393 clusters} \\
Jev 1.13.0 & 88.15 & 87.59 & -0.56 [-1.17, +0.07] \\
Jev 1.14.0 & 87.65 & 87.30 & -0.35 [-1.03, +0.35] \\
Gemma-4-31B & 88.84 & 88.05 & -0.80 [-1.56, -0.05] \\
Jais-2-8B & 81.18 & 80.02 & -1.16 [-2.15, -0.17] \\
\addlinespace
\multicolumn{4}{@{}l}{\textbf{DialectalArabicMMLU}: 15{,}675 pairs; 3{,}135 clusters} \\
Jev 1.13.0 & 85.68 & 81.10 & -4.58 [-5.30, -3.88] \\
Jev 1.14.0 & 84.75 & 81.12 & -3.63 [-4.31, -2.95] \\
Gemma-4-31B & 78.98 & 75.53 & -3.44 [-4.24, -2.63] \\
Jais-2-8B & 53.24 & 49.40 & -3.84 [-4.92, -2.75] \\
\addlinespace
\multicolumn{4}{@{}l}{\textbf{ArBanking77 (Palestinian)}: 3{,}073 pairs; 3{,}073 clusters} \\
Jev 1.13.0 & 75.30 & 66.81 & -8.49 [-9.83, -7.19] \\
Jev 1.14.0 & 74.85 & 66.91 & -7.94 [-9.27, -6.64] \\
Gemma-4-31B & 75.87 & 71.48 & -4.39 [-5.48, -3.30] \\
Jais-2-8B & 20.37 & 16.56 & -3.81 [-4.73, -2.88] \\
\addlinespace
\bottomrule
\end{tabularx}

\caption{Matched accuracy for the fixed RQ2 model panel. Gaps are dialect minus MSA in percentage points, with pointwise 95\% whole-cluster percentile-bootstrap intervals. Pairs count source--variety comparisons; repeated MSA sources and paraphrases remain linked in resampling. ArBanking77 uses source-level means of paraphrase correctness. All four models use identical eligible items.}
\label{tab:rq2-paired-gaps}
\end{table}

\begin{figure}[!htbp]
\centering
\includegraphics[width=\linewidth]{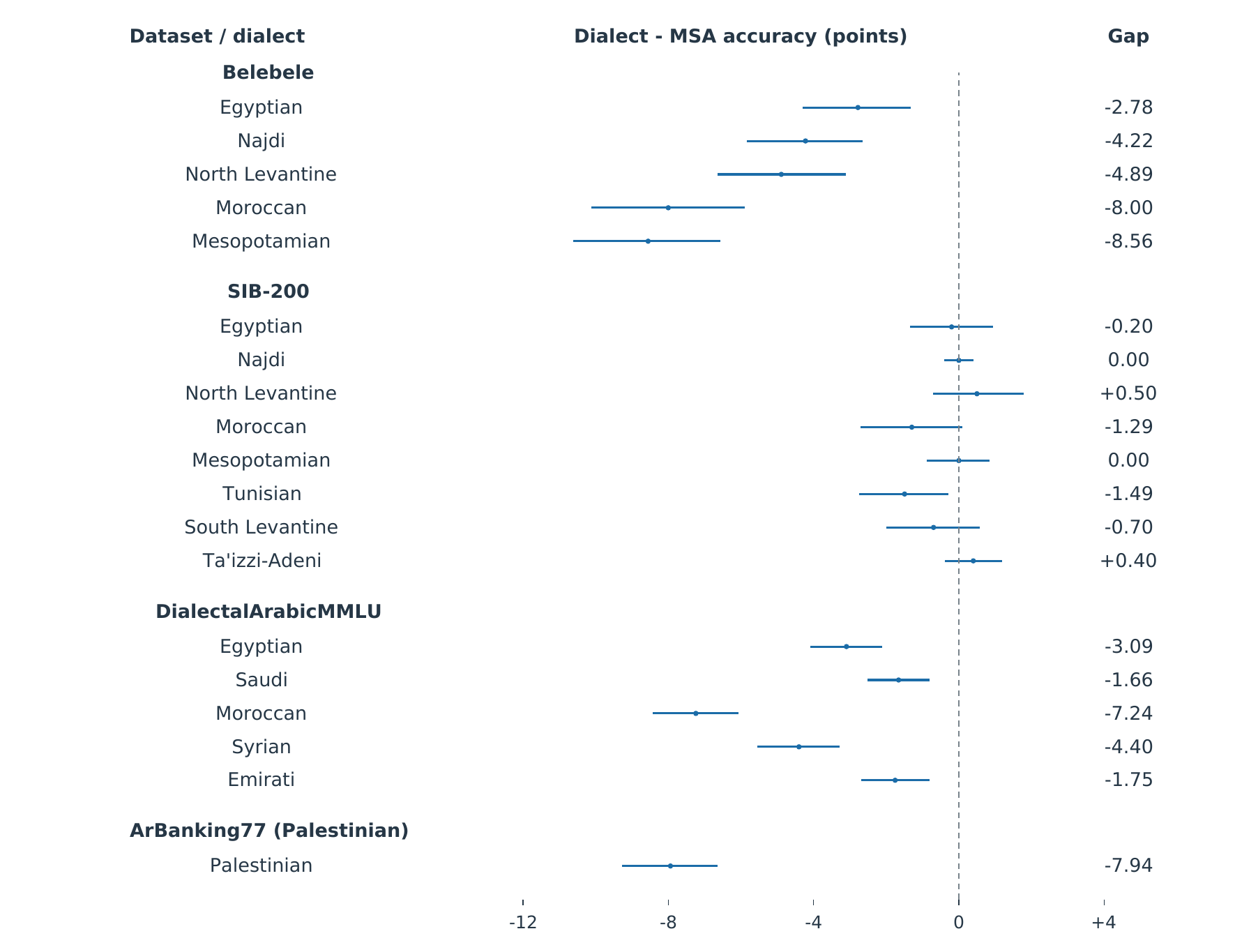}
\caption{Jev~1.14.0's matched dialect-minus-MSA accuracy gap for every retained dataset--dialect cell. Bars are pointwise 95\% source/article-clustered intervals. The zero line denotes equal matched accuracy, not a tested equivalence margin. The main table pools these cells within each dataset; it does not pool datasets.}
\label{fig:rq2-dialects}
\end{figure}

\paragraph{Variation between dialects.}
The pooled estimates conceal substantial differences (\cref{fig:rq2-dialects}). On Belebele, Jev~1.14.0's point gaps range from \RQTwoEgyptBelebeleGap{} points for Egyptian to \RQTwoMesoBelebeleGap{} for Mesopotamian Arabic. On DialectalArabicMMLU, the corresponding Saudi and Moroccan gaps are \RQTwoSaudiMMLUGap{} and \RQTwoMoroccoMMLUGap{} points. These are descriptive within-dataset patterns, not a ranking of intrinsic dialect difficulty: varieties also differ in wording and translation quality. \Cref{tab:rq2-dialect-detail} reports every dialect's absolute accuracy and gap for all four models.

\paragraph{Comparison with other systems.}
The pooled cross-system contrasts vary by dataset (\cref{tab:rq2-contrasts}). On Palestinian ArBanking77, the Jev~1.14.0-minus-Gemma gap contrast is \RQTwoBankGemmaContrast{} points and the Jev-minus-Jais contrast is \RQTwoBankJaisContrast{} points. Both simultaneous bands lie below zero, indicating a larger penalty for Jev on this population. All contrasts outside these two Palestinian banking comparisons, whose single-dialect and pooled rows coincide, have simultaneous bands that include zero (\cref{tab:rq2-contrast-detail}). This does not establish equal penalties. Jais's smaller banking gap accompanies only \RQTwoBankJaisMSA{}\% MSA accuracy and \RQTwoBankJaisDialect{}\% dialect accuracy, illustrating why a gap alone is insufficient to rank useful performance. The stronger absolute banking accuracy for Gemma and its smaller observed gap provide a more direct contrast with Jev on this matched population.

\begin{table}[!htbp]
\centering\small
\begin{tabularx}{\linewidth}{@{}Xrrr@{}}
\toprule
Dataset & vs. Jev 1.13 & vs. Gemma & vs. Jais \\
\midrule
Belebele & +0.42 & -0.80 & +3.22 \\
 & [-1.02, +1.86] & [-2.82, +1.22] & [-0.22, +6.66] \\
\addlinespace
SIB-200 & +0.21 & +0.45 & +0.81 \\
 & [-1.15, +1.57] & [-1.11, +2.00] & [-1.36, +2.98] \\
\addlinespace
DialectalArabicMMLU & +0.95 & -0.19 & +0.21 \\
 & [-0.33, +2.23] & [-1.87, +1.50] & [-1.97, +2.39] \\
\addlinespace
ArBanking77 (Palestinian) & +0.55 & -3.55 & -4.13 \\
 & [-1.06, +2.17] & [-5.60, -1.50] & [-6.34, -1.92] \\
\addlinespace
\bottomrule
\end{tabularx}

\caption{Pooled gap contrasts: Jev~1.14.0 minus each comparator, in percentage points. The second line gives approximate simultaneous 95\% bands accounting for all \RQTwoContrastCount{} dialect-specific and pooled contrasts across four datasets. Positive values mean a less negative gap for Jev. Gemma is Gemma-4-31B; Jais is Jais-2-8B. Bands spanning zero do not establish equivalence.}
\label{tab:rq2-contrasts}
\end{table}

\paragraph{Version changes and identical-input sensitivity.}
Jev~1.14.0's pooled gap is less negative than Jev~1.13.0's in each dataset, but none of the version contrasts is separated from zero by the simultaneous bands. Moreover, a smaller gap need not mean better dialect accuracy. On DialectalArabicMMLU, dialect accuracy is nearly unchanged (\RQTwoMMLUOldDialect{}\% to \RQTwoMMLUNewDialect{}\%), while MSA accuracy falls from \RQTwoMMLUOldMSA{}\% to \RQTwoMMLUNewMSA{}\%. Removing pairs whose complete model inputs are identical across language versions leaves Jev's pooled gap on SIB-200 small, at \RQTwoSIBNonidenticalGap{} points on \RQTwoSIBNonidenticalPairs{} pairs; its pointwise interval still includes zero. \Cref{tab:rq2-nonidentical} reports this sensitivity for all four datasets.

\paragraph{Answer to RQ2.}
On the evaluated matched resources, Jev~1.14.0's dialect penalty depends on both dataset and variety. The point gaps are generally negative on reading comprehension, knowledge questions and Palestinian banking intents, while the topic-classification average is close to zero. The cross-system evidence does not support a general dialect advantage for Jev: it has a larger observed Palestinian banking penalty than both selected LLMs, while the other contrasts remain inconclusive under simultaneous uncertainty. These comparisons concern the particular matched versions and interfaces; they do not isolate task type as the cause, establish native-text performance or measure linguistic distance from MSA.

\section{Discussion and Limitations}\label{sec:discussion}
\subsection{Discussion}
Jev~1.14.0 improves the mean task-primary score over Jev~1.13.0, but both the leading LLM and the leading non-Jev decision model score higher overall in this inventory (\cref{sec:rq1}). Its position also changes across families: the overall increase between Jev versions includes lower knowledge and rating/grading means. The system order also survives removal of the recorded English controls and repeated MMMLU content, although task means remain conditional on coverage. For an application, the relevant task family and supported inputs therefore provide information that a single aggregate rank cannot supply. These comparisons describe the selected systems and protocol; they do not isolate a training or architecture effect.

The matched comparisons show that the language-version effect differs across datasets and varieties. Jev~1.14.0's smaller pooled SIB-200 gap persists after identical-input pairs are removed, while larger losses appear for some varieties on Belebele and DialectalArabicMMLU. Cross-system contrasts add a practical qualification: on matched Palestinian banking intents, Jev loses more accuracy between MSA and dialect versions than either selected LLM. A smaller gap alone is insufficient for choosing a model, as Jais's low banking accuracy shows. Likewise, the smaller gap in a newer Jev version can reflect lower MSA accuracy rather than an improvement on dialect text.

Holding source content fixed helps define the comparison, but datasets still differ in task, wording and included varieties. The present results therefore do not identify task complexity, model architecture or linguistic distance as the cause of a gap. Native text is a separate validation need. The companion supplies a prospective, unexecuted protocol for Egyptian and Gulf sentiment tweets with human-checked MSA and same-dialect paraphrases. The paraphrase condition would help distinguish MSA normalization from rewriting in general. Human review and new-condition evaluation are required before this can supply evidence; the current paper makes no empirical native-text robustness claim.

\subsection{Limitations}

The suite covers selected Arabic tasks and varieties rather than a representative sample of all Arabic use.
Its task average gives equal weight to datasets with different primary metrics, sizes and purposes, so it summarizes the chosen inventory rather than a general level of Arabic ability.
Label definitions also constrain interpretation: identical inputs can have different reference labels in the MADAR city task, as described in \cref{sec:data-constraints}.
For a system that returns the same answer to identical inputs, some errors then reflect conflicting reference labels rather than distinguishable text differences.

The comparison concerns the specified interfaces and inference conditions.
Open LLMs use restricted answer codes with thinking disabled; decision models use their supported native interfaces or adapters.
These conditions allow common task scoring, but do not isolate architecture or training as the cause of a difference.
They also do not establish how the systems would compare with task-specific prompt tuning or generated reasoning.
The primary comparison fixes one task-template/readout condition and does not establish robustness to other prompts or answer representations.

Incomplete support can change the population behind a reported score.
Coverage of the designated primary question does not certify completion of every auxiliary question, and an undefined metric is distinct from a missing response.
Paired comparisons further restrict analysis to the items answered under both conditions.
Different models, tokenizers and subset scopes can therefore contribute different items even when the nominal task is the same.
Coverage and eligibility qualify the comparisons; they do not recover performance on unsupported inputs.

Shared content limits the independence of some observations.
Parallel variety versions share source material, and the MMMLU Arabic and DialectalArabicMMLU overlap remains in the registered task summaries; the separate sensitivity analysis addresses that overlap descriptively.
The exclusions used for the model-card comparisons cover documented overlap lists, while QARAR's unpublished training corpus prevents a complete check of its training exposure.
More generally, the present evaluation cannot establish that public benchmark material was absent from every evaluated model's training data.
We therefore do not interpret held-out evaluation splits as proof of unseen training content.

Primary-score reconstruction and the new source-aware matched contrasts address the numerical checks needed for the two descriptive questions. They do not repair the original RQ1/RQ2 hypothesis specifications or validate source annotations and historical execution settings. The fixed comparison panel represents selected systems, not every model in the inventory, and its simultaneous bands are approximate bootstrap summaries. Those boundaries are recorded in \cref{app:paper-scope}; descriptive results do not replace confirmatory tests.

\section{Conclusion}\label{sec:conclusion}
We evaluated Jev alongside decision models and LLMs on the selected Arabic-focused frozen task suite, then compared matched MSA--dialect performance by variety and against a fixed model panel. Jev~1.14.0 improves the overall task mean over Jev~1.13.0, while its relative performance and version gains vary across families; it does not lead the aggregate comparison. Matched dialect gaps depend on dataset and variety. SIB-200 has a small pooled gap, whereas Palestinian banking data show a larger gap for Jev than for the selected LLMs under simultaneous uncertainty. The evidence supports reporting absolute performance, coverage and matched changes together. It does not establish a general dialect advantage, native-text robustness or the cause of the observed differences.

\section*{Ethics and Artifact Availability}
The evaluation uses existing datasets and their supplied reference labels, including harmful-language, cultural and educational judgments. The mappings and exclusions in \cref{app:mapping-details} define the measured decisions; agreement with these labels does not establish unbiased or safe behavior outside the evaluated conditions. We report task/system aggregates and use an authored example to explain the interface, rather than reproducing source tweets in the template examples.

Evaluation response archives are available from the study's Zenodo record at \url{https://doi.org/10.5281/zenodo.23954443}. The appendices provide the task and system inventories, key mapping choices and reporting corrections. Companion files retain full revisions, detailed mappings and machine-readable results. Access to underlying datasets follows the source-specific terms and controls; source identification does not grant redistribution permission (\cref{app:task-source-revisions}). The preregistration deposit is identified by \url{https://doi.org/10.5281/zenodo.23586552}; its deposited access record specifies an embargo. Dataset texts and the full response archive are not bundled with this manuscript, and we make no claim that the preregistration deposit's files are publicly downloadable.
\section*{Acknowledgements}
This research has been partially funded by the Federal Ministry of Education and Research of Germany and the state of North-Rhine Westphalia as part of the Lamarr-Institute for Machine Learning and Artificial Intelligence.

\bibliographystyle{amlpreprint}
\bibliography{style_controls,verified,benchmark_sources,related_sources,methods_sources,task_sources,task_sources_040,task_sources_041,task_sources_042,task_sources_043,task_sources_044,task_sources_045,task_sources_046,metric_sources_084,publication_sources_090}
\appendix
\section{Benchmark and Evaluation Details}\label{app:supporting}
\subsection{Tasks and Sources}\label{app:task-inventory}
\Cref{tab:task-inventory} identifies all tasks, their question types, primary metrics, frozen sizes and source resources. The primary metric selects one decision or combines the required binary heads; auxiliary questions do not enter the overall mean. Complete parser and sampling descriptions are supplied in the companion file \texttt{task\_mappings.md}.
\begingroup\small
\begin{longtable}{@{}lp{.31\linewidth}lrrp{.13\linewidth}@{}}
\caption{Task inventory and source attribution. Types are Choice (C), Noul (N) and Score (S); trailing digits count questions. Items gives the frozen task size, not independent texts or completed answers. Full mappings and revisions are in the companion files.}\label{tab:task-inventory}\\
\toprule
Family & Task & Types & Metric & Items & Source \\
\midrule
\endfirsthead
\toprule
Family & Task & Types & Metric & Items & Source \\
\midrule
\endhead
\bottomrule
\endfoot
A & Belebele & C & Acc. & 7{,}200 & \citep{bandarkar-etal-2024-belebele} \\
A & SIB-200 & C & Acc. & 11{,}044 & \citep{adelani-etal-2024-sib} \\
A & DialectalArabicMMLU & C & Acc. & 21{,}945 & \citep{altakrori-etal-2026-dialectalarabicmmlu} \\
A & TUNIZI & C & Acc. & 3{,}000 & \citep{fourati-etal-tunizi-resource} \\
B & NADI 2021 & C2 & Macro-F1 & 5{,}000 & \citep{abdul-mageed-etal-2021-nadi} \\
B & ArSarcasm-v2 & N, C2 & Macro-F1 & 3{,}000 & \citep{abu-farha-etal-2021-overview} \\
B & Mawqif & C2, N & Macro-F1 & 619 & \citep{alturayeif-etal-2022-mawqif} \\
C & MADAR-26 & N, C3 & Acc. & 5{,}200 & \citep{bouamor-etal-2018-madar} \\
C & LABR & C3, S & Spearman & 5{,}000 & \citep{aly-atiya-2013-labr} \\
C & SemEval-2018 Affect & N11, S4 & Jaccard & 1{,}518 & \citep{mohammad-etal-2018-semeval} \\
C & MASSIVE ar-SA & C2 & Acc. & 2{,}974 & \citep{fitzgerald-etal-2023-massive} \\
C & SemEval-2016 ABSA hotels & C & Macro-F1 & 300 & \citep{pontiki-etal-2016-semeval} \\
D & ArBanking77 & C & Macro-F1 & 15{,}534 & \citep{jarrar-etal-2023-arbanking77,malaysha-etal-2024-arafinnlp} \\
D & XNLI Arabic & C & Acc. & 5{,}010 & \citep{conneau-etal-2018-xnli} \\
D & MIRACL Arabic & N & F1 & 3{,}000 & \citep{zhang-etal-2023-miracl} \\
D & SemEval-2017 STS & S & Spearman & 250 & \citep{cer-etal-2017-semeval} \\
D & Egyptian offensive and hate speech & N, C & Macro-F1 & 1{,}634 & \citep{ahmed-etal-2022-egyptian} \\
D & AraTrust & C & Acc. & 506 & \citep{alghamdi-etal-2024-aratrust} \\
D & OSACT 2020 offensive & N, C & Macro-F1 & 2{,}000 & \citep{mubarak-etal-2020-overview} \\
D & Arabic spam and ham tweets & N & F1 & 300 & \citep{kaddoura-henno-2024-spam-data,kaddoura-henno-2024-spam} \\
D & ArCyC cyberbullying & N2 & F1 & 300 & \citep{shannag-2023-arcyc} \\
D & AraTox toxicity & N6 & Jaccard & 300 & \citep{alshargi-etal-2026-aratox} \\
D & Multilingual HateCheck & N & F1 & 300 & \citep{rottger-etal-2022-multilingual} \\
D & Aya red-teaming & N9 & Jaccard & 300 & \citep{coherelabs-aya-redteaming-card} \\
E & ArabicMMLU & C & Acc. & 14{,}455 & \citep{koto-etal-2024-arabicmmlu} \\
E & MMMLU Arabic & C & Acc. & 14{,}042 & \citep{openai-mmmlu} \\
F & CheckThat! 2025 subjectivity & N & F1 & 300 & \citep{ruggeri-etal-2025-subjectivity} \\
F & ArMPro propaganda & N & F1 & 300 & \citep{hasanain-etal-2024-gpt} \\
F & CheckThat! 2022 tweets & N2 & F1 & 300 & \citep{nakov-etal-2022-relevant} \\
F & ArAIEval 2023 disinformation & N & F1 & 300 & \citep{hasanain-etal-2023-araieval} \\
F & AIRABIC & N & F1 & 300 & \citep{airabic-maintainer-resource,alshammari-elsayed-2023-airabic} \\
F & HalluTruthQA-4K & N & F1 & 300 & \citep{bekhouche-hallutruthqa-card} \\
G & ArabCulture & C & Acc. & 3{,}458 & \citep{sadallah-etal-2025-commonsense} \\
G & Arabic ComVE & C & Acc. & 1{,}000 & \citep{tawalbeh-al-smadi-2020-valid} \\
G & Arabic Humor & N & F1 & 300 & \citep{alkhalifa-etal-2022-dataset} \\
G & Global PIQA & C & Acc. & 300 & \citep{mrl-global-piqa-nonparallel-card} \\
H & BAREC 2025 readability & S & Spearman & 300 & \citep{camel-lab-barec-sent} \\
H & AR-AES essay scoring & S & Spearman & 300 & \citep{ghazawi-simpson-2024-essay} \\
H & AR-ASAG short-answer grading & S & Spearman & 300 & \citep{ouahrani-bennouar-2020-arasag-data} \\
H & CARE & S & Spearman & 300 & \citep{guo-etal-2025-care} \\
I & Poetry meter (MetRec) & C & Macro-F1 & 300 & \citep{al-shaibani-etal-2020-metrec,al-shaibani-etal-2020-meter} \\
I & Word sense disambiguation & C & Acc. & 300 & \citep{kaddoura-nassar-2024-wsd-data,kaddoura-nassar-2024-wsd} \\
I & Nahw-MCQ grammar & C & Acc. & 300 & \citep{mubarak-etal-2026-nahw} \\
\end{longtable}

\endgroup

\subsection{Mapping Choices and Source Limits}\label{app:mapping-details}\label{app:task-source-revisions}
\paragraph{Parallel and independently collected text.}
Belebele, SIB-200 and DialectalArabicMMLU retain shared source identifiers across language versions. Their frozen primary inventories also include English controls; SIB-200 combines the source train, validation and test splits without fitting a model. English and romanized-MSA controls do not enter the matched dialect comparison. ArBanking77's matched analysis uses corpus-linked MSA and Palestinian sources, averaging available paraphrase correctness within each source and variety. Independently collected variety text, including Global PIQA, does not provide matched MSA--dialect pairs.

\paragraph{Primary targets.}
ArSarcasm-v2 is scored on sentiment even though requests also ask about sarcasm and variety. NADI uses country-level macro-F1; MADAR uses city accuracy; MASSIVE uses intent accuracy. MIRACL becomes a query--passage relevance decision, rather than a retrieval-ranking task. Egyptian harmful-language evaluation uses category macro-F1, with offensive text included in the auxiliary harmfulness label. AraTox scores six toxicity types, excluding the complementary non-toxic label. The CheckThat! 2025 adaptation uses positive-class F1 rather than the source task's macro-F1, so its numbers are not directly comparable with that leaderboard\citep{ruggeri-etal-2025-subjectivity}.

\paragraph{Ordered scores.}
BAREC uses seven levels, mapped to indices 0--6; its nineteen-level scheme exceeds Jev's supported Score range. LABR uses star rating minus one. AR-AES retains rubric maxima of five and averages the available final rater scores. AR-ASAG uses marks on levels 0--5. CARE maps the supplied 1--10 ratings to indices 0--9. Spearman correlation evaluates rank association with these supplied scores, not exact grade agreement or deployment quality.

\paragraph{Sampling and development.}
Added tasks use the deterministic, stratified sampling and duplicate/conflict filtering described in \cref{sec:evaluation-items}; original tasks retain their own rules. NADI uses its source development split for evaluation. MASSIVE uses Arabic release 1.1; its publication describes 1.0, and the maintainers report that the other language data were unchanged when Catalan was added\footnote{\url{https://github.com/alexa/massive}, release notes accessed 2 November 2026.}. Missing auxiliary gold labels, such as emotion intensity, differ from missing required primary answers. The companion mapping descriptions record each exception and development/pilot rule.

\paragraph{Revisions and permissions.}
\texttt{task\_sources.csv} records the complete declared source locators and revisions. These identify intended inputs rather than certifying every source annotation, training exclusion or redistribution right. Historical terms are distinguished from current maintainer labels. In particular, Arabic ComVE has a CC0 licence file and a CC BY-SA statement in its README\footnote{\url{https://github.com/msmadi/Arabic-Dataset-for-Commonsense-Validationion}, accessed 4 November 2026.}; the companion records retain this conflict. Neither dataset texts nor full model responses are bundled with the paper.

\subsection{Systems and Primary Readouts}\label{app:model-inventory}
\Cref{tab:active-model-inventory} lists every evaluated version. The companion \texttt{systems.csv} gives full repository identifiers, pinned revisions, declared precision and adapter settings; \texttt{deferred\_systems.csv} identifies the four excluded systems. Declared precision and service labels do not independently establish historical execution settings or hosted weights. The analysis environment and available run provenance are recorded separately in \texttt{reproducibility.md}.

\begingroup\small
\begin{longtable}{@{}p{.51\linewidth}ll@{}}
\caption{Evaluated systems and declared primary readouts. API: hosted interface; TC: restricted token codes; NH: native heads; OI: own interface; LP: letter codes with its own prompt. Full repository identifiers,revisions and declared precision are in \texttt{systems.csv}.}\label{tab:active-model-inventory}\\
\toprule
System & Declared precision & Readout \\
\midrule
\endfirsthead
\toprule
System & Declared precision & Readout \\
\midrule
\endhead
\bottomrule
\endfoot
\multicolumn{3}{@{}l}{\textbf{Hosted decision models}}\\
jev-1.13.0 & Not recorded & API \\
jev-1.14.0 & Not recorded & API \\
Microsoft-Decision-1 & Not recorded & API \\
OpenAI-Decisions-gpt-6-luna & Not recorded & API \\
\multicolumn{3}{@{}l}{\textbf{Arabic decision models}}\\
QARAR & float32 & NH \\
jeb & float32 & NH \\
Jev-AR & float32 & NH \\
jeb-typed-decisions & float32 & NH \\
ara-jev-Faisal & bfloat16 & LP \\
\multicolumn{3}{@{}l}{\textbf{Other general decision models}}\\
Laya-multilingual & float16 & NH \\
Kev-4B & bfloat16 & NH \\
Kev-27B & bfloat16 & NH \\
SII-Decision-4B & bfloat16 & NH \\
SII-Decision-27B & bfloat16 & NH \\
\multicolumn{3}{@{}l}{\textbf{Arabic-centric LLMs}}\\
Jais-family-256M & float32 & TC \\
Jais-family-590M & float32 & TC \\
Jais-family-1.3B & float32 & TC \\
Jais-family-2.7B & float32 & TC \\
Jais-family-6.7B & float32 & TC \\
Jais-family-13B & float32 & TC \\
Jais-2-8B & bfloat16 & TC \\
ALLaM-7B & bfloat16 & TC \\
Fanar-2-27B & bfloat16 & TC \\
\multicolumn{3}{@{}l}{\textbf{Qwen LLMs}}\\
Qwen3.5-0.8B & bfloat16 & TC \\
Qwen3.5-2B & bfloat16 & TC \\
Qwen3.5-4B & bfloat16 & TC \\
Qwen3.5-9B & bfloat16 & TC \\
Qwen3.5-27B & bfloat16 & TC \\
Qwen3.8-27B & bfloat16 & TC \\
\multicolumn{3}{@{}l}{\textbf{Gemma LLMs}}\\
Gemma-4-E2B & bfloat16 & TC \\
Gemma-4-E4B & bfloat16 & TC \\
Gemma-4-12B & bfloat16 & TC \\
Gemma-4-26B-A4B & bfloat16 & TC \\
Gemma-4-31B & bfloat16 & TC \\
\multicolumn{3}{@{}l}{\textbf{Additional vendor decision models}}\\
pplx-decider-v1.2 & bfloat16 & OI \\
pplx-decider-flash-v1.2 & bfloat16 & OI \\
GLiDE-1.2 & bfloat16 & OI \\
GLiNER2.5-Decide & float32 & OI \\
GLiNER2.5-Decide-1B & float32 & OI \\
GLiNER2.5-multi-Decide & float32 & OI \\
clef & bfloat16 & OI \\
clef-flash & bfloat16 & OI \\
h2o-lightning-31b & bfloat16 & OI \\
h2o-lightning-4b-nemotron-nano & bfloat16 & OI \\
Intern-Decision-4B & bfloat16 & OI \\
Intern-Decision-2B & bfloat16 & OI \\
Intern-Decision-0.8B & bfloat16 & OI \\
d1-3B & bfloat16 & OI \\
d1-omni-600M & float32 & OI \\
tasksource-decider-nano & bfloat16 & OI \\
strands-decider-2B-qwen3.5 & bfloat16 & OI \\
strands-decider-E2B-gemma4 & bfloat16 & OI \\
strands-decider-E4B-gemma4 & bfloat16 & OI \\
strands-decider-12B-gemma4 & bfloat16 & OI \\
strands-decider-26B-A4B-gemma4 & bfloat16 & OI \\
d3 & bfloat16 & OI \\
d3-flash & bfloat16 & OI \\
d3-mini & bfloat16 & OI \\
d3-nano & bfloat16 & OI \\
d3-lite & bfloat16 & OI \\
d3-edge & bfloat16 & OI \\
Vela-2.0-0.3B & float32 & OI \\
Vela-2.0-0.8B & bfloat16 & OI \\
Vela-2.0-4B & bfloat16 & OI \\
Vela-2.0-9B & bfloat16 & OI \\
Laya & float16 & NH \\
Laya-typed-decisions & float16 & NH \\
\end{longtable}

\endgroup

\subsection{Per-Request Timing and Processing Rates}\label{app:runtime-provenance}
We match saved response timings to the exact frozen T-EN requests used in the primary audit, retaining one response record per unique request. A record contains a shared state and its question set; it can include unsupported or skipped questions, so primary coverage remains a separate result. The companion \texttt{request\_timing.csv} reports mean seconds per record and Records/s beside each system's evaluation score. Records/s equals the number of timed records divided by the sum of their recorded processing times, or equivalently the reciprocal of mean seconds per record. Timing is averaged over distinct records, while evaluation scores remain unweighted means over available tasks.

Jev and other decision backends time each request or prediction call, including any network, retry and adapter work. Local LLMs allocate batch processing time across records. Model startup, dataset loading and output-file writes are excluded. The resulting rates describe each backend's recorded execution settings; hardware, batching and concurrency differ across systems. The companion retains the full timing definitions, quantiles, counts and source hashes.

\subsection{Frozen Primary Wording}\label{app:frozen-templates}
The T-EN source files match the recorded frozen hashes. \Cref{tab:frozen-template-example} shows an actual task question without an evaluation tweet or model prediction. Task wording is separate from the LLM wrapper, which begins ``Answer the question about the following state.'' and adds answer-format instructions. The default readout normalizes the probabilities of supported single-token answer codes; context limits and unsupported representations affect coverage. The companion \texttt{templates\_primary.json} retains the exact primary source strings, not each backend's final chat serialization.
\begingroup\small
\begin{longtable}{@{}p{.16\linewidth}p{.78\linewidth}@{}}
\caption{Actual frozen ArSarcasm-v2 sentiment question under the primary T-EN template. Internal label keys are preserved; no tweet, predicted answer or full model-specific chat serialization is shown. Exact source strings for the complete main suite are retained in the accompanying local template bundle.}\label{tab:frozen-template-example}\\
\toprule
Template & Frozen instruction and criterion descriptions \\
\midrule
\endfirsthead
\toprule
Template & Frozen instruction and criterion descriptions \\
\midrule
\endhead
\bottomrule
\endfoot
T-EN & \textbf{Instruction:} What is the overall sentiment of `tweet`?\par\texttt{negative}: The text expresses a negative opinion.\par\texttt{neutral}: The text expresses neither a positive nor a negative opinion.\par\texttt{positive}: The text expresses a positive opinion. \\[.8ex]
\end{longtable}

\endgroup

\section{Additional Descriptive Results}
\subsection{All-System Summaries and Exclusion Sensitivities}\label{app:language-overlap-sensitivity}\label{app:complete-results}
\Cref{tab:all-system-results-timing} reports every system's frozen evaluation score and recorded processing rate. The companion \texttt{sensitivities.csv} retains all three descriptive masks. The Arabic-only mask removes 900 English items from Belebele, 1,004 from SIB-200 and 3,135 from DialectalArabicMMLU, while retaining romanized MSA and T-EN instructions. The overlap mask removes 2,391 MMMLU items whose state/question requests also occur in DialectalArabicMMLU; it removes them only from MMMLU. Each affected task metric is recomputed before unweighted task averaging. The combined mask applies both exclusions. All 67 systems retain their ordering; maximum absolute mean changes are below \ArabicMaxChangeBound{} and \RawOverlapChangeBound{} aggregate points for the separate masks, respectively.

The full unrounded system--task matrix, item counts and coverage are supplied as \texttt{primary\_scores.csv}. Its original export is retained separately as \texttt{primary\_scores\_original.csv}, with the single reporting correction documented below. These machine-readable tables replace an exhaustive printed score matrix.
\begingroup\small
\begin{longtable}{@{}p{.44\linewidth}rrrr@{}}
\caption{All systems ordered by the unrounded evaluation mean. Scores are mean task-primary metrics multiplied by 100. Records/s uses recorded processing time under each backend's execution settings (\cref{app:runtime-provenance}). Tasks and coverage describe the scored population. Masked means are in the companion \texttt{sensitivities.csv}.}\label{tab:all-system-results-timing}\\
\toprule
System & Score & Records/s & Tasks & Coverage (\%) \\
\midrule
\endfirsthead
\toprule
System & Score & Records/s & Tasks & Coverage (\%) \\
\midrule
\endhead
\bottomrule
\endfoot
Gemma-\allowbreak{}4-\allowbreak{}31B & 71.81 & 7.86 & 43/43 & 100.00 \\
h2o-\allowbreak{}lightning-\allowbreak{}31b & 71.12 & 7.11 & 43/43 & 100.00 \\
d3 & 70.78 & 4.05 & 43/43 & 100.00 \\
GLiDE-\allowbreak{}1.2 & 70.36 & 4.23 & 43/43 & 100.00 \\
pplx-\allowbreak{}decider-\allowbreak{}v1.2 & 69.90 & 4.88 & 43/43 & 100.00 \\
jev-\allowbreak{}1.14.0 & 69.24 & 4.93 & 43/43 & 100.00 \\
OpenAI-\allowbreak{}Decisions-\allowbreak{}gpt-\allowbreak{}6-\allowbreak{}luna & 68.79 & 3.57 & 43/43 & 99.84 \\
jev-\allowbreak{}1.13.0 & 68.77 & 5.22 & 43/43 & 100.00 \\
Gemma-\allowbreak{}4-\allowbreak{}26B-\allowbreak{}A4B & 68.55 & 26.56 & 43/43 & 100.00 \\
SII-\allowbreak{}Decision-\allowbreak{}27B & 68.39 & 2.06 & 43/43 & 100.00 \\
Kev-\allowbreak{}27B & 67.51 & 4.70 & 43/43 & 100.00 \\
d3-\allowbreak{}flash & 67.04 & 9.60 & 43/43 & 100.00 \\
clef & 66.82 & 1.95 & 43/43 & 100.00 \\
strands-\allowbreak{}decider-\allowbreak{}26B-\allowbreak{}A4B-\allowbreak{}gemma4 & 66.70 & 4.06 & 43/43 & 100.00 \\
Gemma-\allowbreak{}4-\allowbreak{}12B & 66.64 & 18.71 & 43/43 & 100.00 \\
Microsoft-\allowbreak{}Decision-\allowbreak{}1 & 66.06 & 2.79 & 43/43 & 100.00 \\
Qwen3.5-\allowbreak{}27B & 65.96 & 6.42 & 43/43 & 100.00 \\
Qwen3.8-\allowbreak{}27B & 65.15 & 6.53 & 43/43 & 100.00 \\
strands-\allowbreak{}decider-\allowbreak{}12B-\allowbreak{}gemma4 & 64.24 & 5.52 & 43/43 & 100.00 \\
d3-\allowbreak{}mini & 64.00 & 9.91 & 43/43 & 100.00 \\
pplx-\allowbreak{}decider-\allowbreak{}flash-\allowbreak{}v1.2 & 63.49 & 11.13 & 43/43 & 100.00 \\
clef-\allowbreak{}flash & 60.85 & 4.00 & 43/43 & 100.00 \\
strands-\allowbreak{}decider-\allowbreak{}E4B-\allowbreak{}gemma4 & 60.78 & 7.34 & 43/43 & 100.00 \\
SII-\allowbreak{}Decision-\allowbreak{}4B & 60.39 & 4.56 & 43/43 & 100.00 \\
ara-\allowbreak{}jev-\allowbreak{}Faisal & 59.93 & 3.65 & 40/43 & 93.02 \\
Intern-\allowbreak{}Decision-\allowbreak{}4B & 59.75 & 5.27 & 42/43 & 97.67 \\
Qwen3.5-\allowbreak{}9B & 59.43 & 19.43 & 43/43 & 100.00 \\
Gemma-\allowbreak{}4-\allowbreak{}E4B & 59.36 & 36.08 & 43/43 & 100.00 \\
Vela-\allowbreak{}2.0-\allowbreak{}9B & 57.65 & 4.28 & 43/43 & 100.00 \\
Kev-\allowbreak{}4B & 56.87 & 14.28 & 43/43 & 100.00 \\
d1-\allowbreak{}3B & 56.82 & 40.47 & 43/43 & 100.00 \\
d3-\allowbreak{}nano & 56.24 & 14.26 & 43/43 & 100.00 \\
Vela-\allowbreak{}2.0-\allowbreak{}4B & 55.77 & 4.32 & 43/43 & 100.00 \\
Qwen3.5-\allowbreak{}4B & 54.99 & 21.78 & 43/43 & 100.00 \\
Jais-\allowbreak{}2-\allowbreak{}8B & 54.54 & 36.68 & 43/43 & 100.00 \\
ALLaM-\allowbreak{}7B & 54.23 & 32.67 & 43/43 & 100.00 \\
Fanar-\allowbreak{}2-\allowbreak{}27B & 53.68 & 9.72 & 43/43 & 100.00 \\
strands-\allowbreak{}decider-\allowbreak{}E2B-\allowbreak{}gemma4 & 53.66 & 8.63 & 43/43 & 100.00 \\
Intern-\allowbreak{}Decision-\allowbreak{}2B & 51.75 & 7.10 & 42/43 & 97.67 \\
d3-\allowbreak{}lite & 50.43 & 14.39 & 43/43 & 100.00 \\
strands-\allowbreak{}decider-\allowbreak{}2B-\allowbreak{}qwen3.5 & 50.37 & 8.94 & 43/43 & 100.00 \\
h2o-\allowbreak{}lightning-\allowbreak{}4b-\allowbreak{}nemotron-\allowbreak{}nano & 49.45 & 22.66 & 43/43 & 100.00 \\
Jais-\allowbreak{}family-\allowbreak{}13B & 46.84 & 1.37 & 43/43 & 99.60 \\
Gemma-\allowbreak{}4-\allowbreak{}E2B & 45.84 & 60.93 & 43/43 & 100.00 \\
Jais-\allowbreak{}family-\allowbreak{}6.7B & 45.62 & 2.56 & 43/43 & 99.60 \\
Qwen3.5-\allowbreak{}2B & 45.61 & 48.23 & 43/43 & 100.00 \\
Vela-\allowbreak{}2.0-\allowbreak{}0.8B & 44.46 & 5.94 & 43/43 & 100.00 \\
Intern-\allowbreak{}Decision-\allowbreak{}0.8B & 43.93 & 6.82 & 42/43 & 97.67 \\
d3-\allowbreak{}edge & 39.96 & 35.91 & 43/43 & 100.00 \\
Jais-\allowbreak{}family-\allowbreak{}2.7B & 39.69 & 6.20 & 43/43 & 99.60 \\
Jev-\allowbreak{}AR & 37.42 & 61.20 & 22/43 & 50.86 \\
Vela-\allowbreak{}2.0-\allowbreak{}0.3B & 35.92 & 63.21 & 43/43 & 100.00 \\
Qwen3.5-\allowbreak{}0.8B & 35.13 & 57.50 & 43/43 & 100.00 \\
jeb & 34.20 & 79.11 & 40/43 & 93.01 \\
jeb-\allowbreak{}typed-\allowbreak{}decisions & 31.80 & 79.05 & 40/43 & 93.01 \\
GLiNER2.5-\allowbreak{}Decide & 31.69 & 23.83 & 43/43 & 99.998 \\
GLiNER2.5-\allowbreak{}Decide-\allowbreak{}1B & 31.41 & 19.22 & 43/43 & 99.998 \\
GLiNER2.5-\allowbreak{}multi-\allowbreak{}Decide & 31.21 & 48.94 & 43/43 & 99.998 \\
Jais-\allowbreak{}family-\allowbreak{}590M & 30.88 & 23.42 & 43/43 & 99.60 \\
Laya-\allowbreak{}multilingual & 30.43 & 55.43 & 43/43 & 100.00 \\
Jais-\allowbreak{}family-\allowbreak{}256M & 29.31 & 43.42 & 43/43 & 99.60 \\
QARAR & 27.07 & 59.08 & 43/43 & 100.00 \\
Jais-\allowbreak{}family-\allowbreak{}1.3B & 26.71 & 11.77 & 43/43 & 99.60 \\
d1-\allowbreak{}omni-\allowbreak{}600M & 25.37 & 39.44 & 43/43 & 100.00 \\
tasksource-\allowbreak{}decider-\allowbreak{}nano & 22.99 & 46.68 & 37/43 & 86.05 \\
Laya-\allowbreak{}typed-\allowbreak{}decisions & 22.04 & 34.55 & 43/43 & 100.00 \\
Laya & 21.76 & 35.09 & 43/43 & 100.00 \\
\end{longtable}

\endgroup

\subsection{Focal-System Task-Family Means}
The exact focal-system means complement the main heatmap. Comparator selection uses overall means rather than selecting a different leading system within each family.
\begin{table}[htbp]
\centering\small
\begin{tabularx}{\linewidth}{@{}Xrrrr@{}}
\toprule
Family & Jev 1.14 & Jev 1.13 & Gemma 31B & H2O 31B \\
\midrule
A Varieties and script & 83.00 & 82.89 & 84.24 & 84.13 \\
B Native dialect text & 48.14 & 48.08 & 53.22 & 52.56 \\
C Label ladders & 52.03 & 51.16 & 55.15 & 55.33 \\
D Application tasks & 74.08 & 73.41 & 77.87 & 78.74 \\
E Knowledge & 81.95 & 82.47 & 80.19 & 80.80 \\
F News and claims & 70.57 & 69.47 & 68.65 & 60.14 \\
G Commonsense and culture & 82.28 & 82.03 & 86.04 & 86.66 \\
H Rating and grading & 73.49 & 73.71 & 73.29 & 74.48 \\
I Arabic language & 47.08 & 46.60 & 57.09 & 58.50 \\
\bottomrule
\end{tabularx}

\caption{Task-family means multiplied by 100. Gemma 31B is Gemma-4-31B; H2O 31B is h2o-lightning-31b. All four systems have full primary-item coverage. Family composition and primary metrics differ.}
\label{tab:rq1-families}
\end{table}

\subsection{Common Task Support}\label{app:support-sensitivity}
The intersection of finite primary scores for all twelve fixed summary systems contains \CommonTaskCount{} tasks, excluding ArBanking77, MADAR and MASSIVE. \Cref{tab:corrected-support} uses the current corrected primary copy. This restriction changes the task mix rather than ensuring identical answered items. Two close comparisons reverse: Jev~1.13.0 versus OpenAI Decisions, and Qwen3.5-27B versus Microsoft Decision-1.
\begin{table}[htbp]
\centering\small
\begin{tabularx}{\linewidth}{@{}Xrr@{}}
\toprule
System & Corrected mean & Corrected shared-task mean \\
\midrule
Gemma-4-31B & 71.81 & 72.92 \\
h2o-lightning-31b & 71.12 & 71.93 \\
jev-1.14.0 & 69.24 & 70.69 \\
OpenAI-Decisions-gpt-6-luna & 68.79 & 69.93 \\
jev-1.13.0 & 68.77 & 70.21 \\
SII-Decision-27B & 68.39 & 69.71 \\
Microsoft-Decision-1 & 66.06 & 67.04 \\
Qwen3.5-27B & 65.96 & 67.34 \\
Qwen3.8-27B & 65.15 & 66.37 \\
ara-jev-Faisal & 59.93 & 59.93 \\
Jais-2-8B & 54.54 & 57.76 \\
Jais-family-13B & 46.84 & 49.54 \\
\bottomrule
\end{tabularx}

\caption{Available-task and common-task means after the completeness correction, multiplied by 100. The common column retains the same 40 tasks; item support can still differ within them.}
\label{tab:corrected-support}
\end{table}

\subsection{Matched Dialect Panel and Uncertainty}\label{app:rq2-extension-method}
The expanded panel uses the same eligible item UIDs across both Jev versions, Gemma-4-31B and Jais-2-8B. Full state/question hashes and source/gold metadata agree across systems. All primary responses are present for the \RQTwoExtensionItems{} eligible rows before source matching. Only sources with both MSA and a given dialect enter that comparison; unmatched sources are counted in the companion support report. The resulting \RQTwoExtensionPairs{} pairs reproduce every original Jev~1.14.0 matched correctness value. All 16 full-task primary score/count/coverage cells also reproduce, and the response-source hashes match the earlier primary audit. Banking source means retain identical paraphrase sets across systems; these means need not be binary, so an exact McNemar test is not applied to them.

For each dataset, whole-cluster multinomial resampling generates \RQTwoExtensionDraws{} draws with NumPy's default PCG64 generator. Seeds are 12900 for Belebele, 12901 for SIB-200, 12902 for DialectalArabicMMLU and 12903 for ArBanking77. Each draw uses one shared cluster multiplicity vector for all models and varieties. Source--variety pairs have equal weight, so resampled clusters contribute all their pairs and the denominator is recomputed on every draw. MSA observations recur against several varieties and remain linked. The 2.5th and 97.5th percentiles give the pointwise intervals.

The simultaneous grid contains three Jev~1.14.0-minus-comparator gap contrasts for every dialect and the pooled row of each dataset, for \RQTwoContrastCount{} reported contrasts in total. The Palestinian-only banking dialect and pooled rows are identical and are retained for consistent reporting. For each contrast, the bootstrap standard deviation supplies a fixed scale. Within each draw we take the largest absolute centered, scaled deviation over that dataset's entire contrast grid. The 98.75th-percentile critical value gives a two-sided band centered on the observed contrast. Allocating error probability 0.05/4 to each dataset yields approximate 95\% simultaneous coverage across the grid by the union bound, even though Belebele and SIB-200 share FLORES content. These are conservative, bootstrap-based descriptive bands, not adjusted outcomes of the original hypothesis tests. The companion retains pointwise contrast intervals as well as the simultaneous bands.

\begingroup\small
\begin{longtable}{@{}llrrrl@{}}
\caption{Per-dialect matched accuracy and pointwise 95\% intervals. Gaps are dialect minus MSA in percentage points; all four models use identical items.}\label{tab:rq2-dialect-detail}\\
\toprule
Dialect & System & Pairs & MSA (\%) & Dialect (\%) & Gap [95\% interval] \\
\midrule
\endfirsthead
\toprule
Dialect & System & Pairs & MSA (\%) & Dialect (\%) & Gap [95\% interval] \\
\midrule
\endhead
\bottomrule
\endfoot
\multicolumn{6}{@{}l}{\textbf{Belebele}} \\
Egyptian & Jev 1.13.0 & 900 & 95.44 & 93.11 & -2.33 [-3.68, -1.01] \\
Egyptian & Jev 1.14.0 & 900 & 95.56 & 92.78 & -2.78 [-4.30, -1.32] \\
Egyptian & Gemma-4-31B & 900 & 94.44 & 92.00 & -2.44 [-4.01, -0.89] \\
Egyptian & Jais-2-8B & 900 & 82.11 & 75.11 & -7.00 [-9.64, -4.43] \\
Najdi & Jev 1.13.0 & 900 & 95.44 & 90.89 & -4.56 [-6.19, -2.99] \\
Najdi & Jev 1.14.0 & 900 & 95.56 & 91.33 & -4.22 [-5.84, -2.65] \\
Najdi & Gemma-4-31B & 900 & 94.44 & 91.33 & -3.11 [-4.77, -1.45] \\
Najdi & Jais-2-8B & 900 & 82.11 & 75.89 & -6.22 [-8.79, -3.63] \\
North Levantine & Jev 1.13.0 & 900 & 95.44 & 88.89 & -6.56 [-8.59, -4.55] \\
North Levantine & Jev 1.14.0 & 900 & 95.56 & 90.67 & -4.89 [-6.65, -3.11] \\
North Levantine & Gemma-4-31B & 900 & 94.44 & 90.00 & -4.44 [-6.43, -2.45] \\
North Levantine & Jais-2-8B & 900 & 82.11 & 76.00 & -6.11 [-8.95, -3.31] \\
Moroccan & Jev 1.13.0 & 900 & 95.44 & 87.89 & -7.56 [-9.65, -5.47] \\
Moroccan & Jev 1.14.0 & 900 & 95.56 & 87.56 & -8.00 [-10.12, -5.90] \\
Moroccan & Gemma-4-31B & 900 & 94.44 & 88.11 & -6.33 [-8.44, -4.31] \\
Moroccan & Jais-2-8B & 900 & 82.11 & 70.00 & -12.11 [-15.05, -9.13] \\
Mesopotamian & Jev 1.13.0 & 900 & 95.44 & 85.89 & -9.56 [-11.84, -7.40] \\
Mesopotamian & Jev 1.14.0 & 900 & 95.56 & 87.00 & -8.56 [-10.63, -6.56] \\
Mesopotamian & Gemma-4-31B & 900 & 94.44 & 86.33 & -8.11 [-10.23, -6.08] \\
Mesopotamian & Jais-2-8B & 900 & 82.11 & 69.00 & -13.11 [-16.02, -10.34] \\
\addlinespace
\multicolumn{6}{@{}l}{\textbf{SIB-200}} \\
Egyptian & Jev 1.13.0 & 1{,}004 & 88.15 & 87.05 & -1.10 [-2.18, 0.00] \\
Egyptian & Jev 1.14.0 & 1{,}004 & 87.65 & 87.45 & -0.20 [-1.34, +0.93] \\
Egyptian & Gemma-4-31B & 1{,}004 & 88.84 & 88.15 & -0.70 [-2.05, +0.62] \\
Egyptian & Jais-2-8B & 1{,}004 & 81.18 & 80.08 & -1.10 [-2.62, +0.41] \\
Najdi & Jev 1.13.0 & 1{,}004 & 88.15 & 87.95 & -0.20 [-0.51, 0.00] \\
Najdi & Jev 1.14.0 & 1{,}004 & 87.65 & 87.65 & 0.00 [-0.40, +0.40] \\
Najdi & Gemma-4-31B & 1{,}004 & 88.84 & 88.65 & -0.20 [-0.61, +0.19] \\
Najdi & Jais-2-8B & 1{,}004 & 81.18 & 80.68 & -0.50 [-1.17, +0.10] \\
North Levantine & Jev 1.13.0 & 1{,}004 & 88.15 & 87.95 & -0.20 [-1.36, +0.94] \\
North Levantine & Jev 1.14.0 & 1{,}004 & 87.65 & 88.15 & +0.50 [-0.71, +1.78] \\
North Levantine & Gemma-4-31B & 1{,}004 & 88.84 & 88.05 & -0.80 [-2.10, +0.51] \\
North Levantine & Jais-2-8B & 1{,}004 & 81.18 & 80.38 & -0.80 [-2.42, +0.86] \\
Moroccan & Jev 1.13.0 & 1{,}004 & 88.15 & 86.75 & -1.39 [-2.73, -0.10] \\
Moroccan & Jev 1.14.0 & 1{,}004 & 87.65 & 86.35 & -1.29 [-2.71, +0.10] \\
Moroccan & Gemma-4-31B & 1{,}004 & 88.84 & 86.95 & -1.89 [-3.44, -0.40] \\
Moroccan & Jais-2-8B & 1{,}004 & 81.18 & 79.28 & -1.89 [-3.88, +0.10] \\
Mesopotamian & Jev 1.13.0 & 1{,}004 & 88.15 & 87.75 & -0.40 [-1.23, +0.41] \\
Mesopotamian & Jev 1.14.0 & 1{,}004 & 87.65 & 87.65 & 0.00 [-0.89, +0.84] \\
Mesopotamian & Gemma-4-31B & 1{,}004 & 88.84 & 88.25 & -0.60 [-1.55, +0.32] \\
Mesopotamian & Jais-2-8B & 1{,}004 & 81.18 & 79.58 & -1.59 [-2.92, -0.29] \\
Tunisian & Jev 1.13.0 & 1{,}004 & 88.15 & 87.55 & -0.60 [-1.76, +0.59] \\
Tunisian & Jev 1.14.0 & 1{,}004 & 87.65 & 86.16 & -1.49 [-2.75, -0.30] \\
Tunisian & Gemma-4-31B & 1{,}004 & 88.84 & 87.85 & -1.00 [-2.21, +0.10] \\
Tunisian & Jais-2-8B & 1{,}004 & 81.18 & 79.08 & -2.09 [-3.68, -0.51] \\
South Levantine & Jev 1.13.0 & 1{,}004 & 88.15 & 87.65 & -0.50 [-1.82, +0.80] \\
South Levantine & Jev 1.14.0 & 1{,}004 & 87.65 & 86.95 & -0.70 [-2.00, +0.58] \\
South Levantine & Gemma-4-31B & 1{,}004 & 88.84 & 88.15 & -0.70 [-2.03, +0.61] \\
South Levantine & Jais-2-8B & 1{,}004 & 81.18 & 79.78 & -1.39 [-3.00, +0.20] \\
Ta'izzi-Adeni & Jev 1.13.0 & 1{,}004 & 88.15 & 88.05 & -0.10 [-0.97, +0.77] \\
Ta'izzi-Adeni & Jev 1.14.0 & 1{,}004 & 87.65 & 88.05 & +0.40 [-0.39, +1.19] \\
Ta'izzi-Adeni & Gemma-4-31B & 1{,}004 & 88.84 & 88.35 & -0.50 [-1.31, +0.30] \\
Ta'izzi-Adeni & Jais-2-8B & 1{,}004 & 81.18 & 81.27 & +0.10 [-1.04, +1.20] \\
\addlinespace
\multicolumn{6}{@{}l}{\textbf{DialectalArabicMMLU}} \\
Egyptian & Jev 1.13.0 & 3{,}135 & 85.68 & 82.33 & -3.35 [-4.27, -2.46] \\
Egyptian & Jev 1.14.0 & 3{,}135 & 84.75 & 81.66 & -3.09 [-4.08, -2.11] \\
Egyptian & Gemma-4-31B & 3{,}135 & 78.98 & 76.08 & -2.90 [-4.02, -1.82] \\
Egyptian & Jais-2-8B & 3{,}135 & 53.24 & 51.39 & -1.85 [-3.25, -0.38] \\
Saudi & Jev 1.13.0 & 3{,}135 & 85.68 & 82.68 & -3.00 [-3.99, -2.07] \\
Saudi & Jev 1.14.0 & 3{,}135 & 84.75 & 83.09 & -1.66 [-2.52, -0.80] \\
Saudi & Gemma-4-31B & 3{,}135 & 78.98 & 77.16 & -1.82 [-2.84, -0.83] \\
Saudi & Jais-2-8B & 3{,}135 & 53.24 & 49.79 & -3.44 [-4.85, -2.07] \\
Moroccan & Jev 1.13.0 & 3{,}135 & 85.68 & 77.16 & -8.52 [-9.76, -7.30] \\
Moroccan & Jev 1.14.0 & 3{,}135 & 84.75 & 77.51 & -7.24 [-8.42, -6.06] \\
Moroccan & Gemma-4-31B & 3{,}135 & 78.98 & 72.63 & -6.35 [-7.62, -5.07] \\
Moroccan & Jais-2-8B & 3{,}135 & 53.24 & 47.37 & -5.87 [-7.40, -4.31] \\
Syrian & Jev 1.13.0 & 3{,}135 & 85.68 & 80.19 & -5.49 [-6.67, -4.31] \\
Syrian & Jev 1.14.0 & 3{,}135 & 84.75 & 80.35 & -4.40 [-5.55, -3.29] \\
Syrian & Gemma-4-31B & 3{,}135 & 78.98 & 74.96 & -4.02 [-5.30, -2.71] \\
Syrian & Jais-2-8B & 3{,}135 & 53.24 & 48.01 & -5.23 [-6.79, -3.64] \\
Emirati & Jev 1.13.0 & 3{,}135 & 85.68 & 83.13 & -2.55 [-3.51, -1.63] \\
Emirati & Jev 1.14.0 & 3{,}135 & 84.75 & 83.00 & -1.75 [-2.68, -0.80] \\
Emirati & Gemma-4-31B & 3{,}135 & 78.98 & 76.84 & -2.14 [-3.19, -1.05] \\
Emirati & Jais-2-8B & 3{,}135 & 53.24 & 50.43 & -2.81 [-4.15, -1.47] \\
\addlinespace
\multicolumn{6}{@{}l}{\textbf{ArBanking77 (Palestinian)}} \\
Palestinian & Jev 1.13.0 & 3{,}073 & 75.30 & 66.81 & -8.49 [-9.83, -7.19] \\
Palestinian & Jev 1.14.0 & 3{,}073 & 74.85 & 66.91 & -7.94 [-9.27, -6.64] \\
Palestinian & Gemma-4-31B & 3{,}073 & 75.87 & 71.48 & -4.39 [-5.48, -3.30] \\
Palestinian & Jais-2-8B & 3{,}073 & 20.37 & 16.56 & -3.81 [-4.73, -2.88] \\
\addlinespace
\end{longtable}

\begin{longtable}{@{}llrl@{}}
\caption{Jev 1.14.0 minus comparator dialect--MSA gap, by dialect. Bands account for the complete 69-contrast grid. Positive values mean a less negative gap for Jev.}\label{tab:rq2-contrast-detail}\\
\toprule
Dialect & Comparator & Contrast & Simultaneous band \\
\midrule
\endfirsthead
\toprule
Dialect & Comparator & Contrast & Simultaneous band \\
\midrule
\endhead
\bottomrule
\endfoot
\multicolumn{4}{@{}l}{\textbf{Belebele}} \\
Egyptian & Jev 1.13.0 & -0.44 & [-2.22, +1.33] \\
Egyptian & Gemma-4-31B & -0.33 & [-3.13, +2.47] \\
Egyptian & Jais-2-8B & +4.22 & [-0.32, +8.76] \\
Najdi & Jev 1.13.0 & +0.33 & [-1.61, +2.28] \\
Najdi & Gemma-4-31B & -1.11 & [-4.05, +1.82] \\
Najdi & Jais-2-8B & +2.00 & [-2.42, +6.42] \\
North Levantine & Jev 1.13.0 & +1.67 & [-0.96, +4.29] \\
North Levantine & Gemma-4-31B & -0.44 & [-3.57, +2.68] \\
North Levantine & Jais-2-8B & +1.22 & [-3.91, +6.35] \\
Moroccan & Jev 1.13.0 & -0.44 & [-3.00, +2.12] \\
Moroccan & Gemma-4-31B & -1.67 & [-5.50, +2.17] \\
Moroccan & Jais-2-8B & +4.11 & [-1.32, +9.55] \\
Mesopotamian & Jev 1.13.0 & +1.00 & [-1.64, +3.64] \\
Mesopotamian & Gemma-4-31B & -0.44 & [-3.66, +2.77] \\
Mesopotamian & Jais-2-8B & +4.56 & [-0.90, +10.01] \\
\addlinespace
\multicolumn{4}{@{}l}{\textbf{SIB-200}} \\
Egyptian & Jev 1.13.0 & +0.90 & [-1.06, +2.85] \\
Egyptian & Gemma-4-31B & +0.50 & [-1.96, +2.96] \\
Egyptian & Jais-2-8B & +0.90 & [-2.57, +4.37] \\
Najdi & Jev 1.13.0 & +0.20 & [-0.29, +0.69] \\
Najdi & Gemma-4-31B & +0.20 & [-0.65, +1.05] \\
Najdi & Jais-2-8B & +0.50 & [-0.87, +1.87] \\
North Levantine & Jev 1.13.0 & +0.70 & [-1.50, +2.89] \\
North Levantine & Gemma-4-31B & +1.29 & [-1.16, +3.75] \\
North Levantine & Jais-2-8B & +1.29 & [-2.52, +5.11] \\
Moroccan & Jev 1.13.0 & +0.10 & [-2.51, +2.71] \\
Moroccan & Gemma-4-31B & +0.60 & [-2.72, +3.92] \\
Moroccan & Jais-2-8B & +0.60 & [-3.18, +4.38] \\
Mesopotamian & Jev 1.13.0 & +0.40 & [-1.40, +2.19] \\
Mesopotamian & Gemma-4-31B & +0.60 & [-1.32, +2.52] \\
Mesopotamian & Jais-2-8B & +1.59 & [-1.18, +4.37] \\
Tunisian & Jev 1.13.0 & -0.90 & [-3.22, +1.43] \\
Tunisian & Gemma-4-31B & -0.50 & [-3.42, +2.42] \\
Tunisian & Jais-2-8B & +0.60 & [-2.83, +4.03] \\
South Levantine & Jev 1.13.0 & -0.20 & [-2.48, +2.08] \\
South Levantine & Gemma-4-31B & 0.00 & [-2.52, +2.52] \\
South Levantine & Jais-2-8B & +0.70 & [-2.90, +4.29] \\
Ta'izzi-Adeni & Jev 1.13.0 & +0.50 & [-1.41, +2.40] \\
Ta'izzi-Adeni & Gemma-4-31B & +0.90 & [-0.98, +2.78] \\
Ta'izzi-Adeni & Jais-2-8B & +0.30 & [-2.10, +2.70] \\
\addlinespace
\multicolumn{4}{@{}l}{\textbf{DialectalArabicMMLU}} \\
Egyptian & Jev 1.13.0 & +0.26 & [-1.46, +1.97] \\
Egyptian & Gemma-4-31B & -0.19 & [-2.48, +2.09] \\
Egyptian & Jais-2-8B & -1.24 & [-4.17, +1.68] \\
Saudi & Jev 1.13.0 & +1.34 & [-0.32, +3.00] \\
Saudi & Gemma-4-31B & +0.16 & [-1.99, +2.31] \\
Saudi & Jais-2-8B & +1.79 & [-1.05, +4.63] \\
Moroccan & Jev 1.13.0 & +1.28 & [-0.70, +3.25] \\
Moroccan & Gemma-4-31B & -0.89 & [-3.54, +1.75] \\
Moroccan & Jais-2-8B & -1.37 & [-4.67, +1.93] \\
Syrian & Jev 1.13.0 & +1.08 & [-0.82, +2.98] \\
Syrian & Gemma-4-31B & -0.38 & [-3.05, +2.28] \\
Syrian & Jais-2-8B & +0.83 & [-2.39, +4.05] \\
Emirati & Jev 1.13.0 & +0.80 & [-0.83, +2.43] \\
Emirati & Gemma-4-31B & +0.38 & [-1.87, +2.63] \\
Emirati & Jais-2-8B & +1.05 & [-1.71, +3.81] \\
\addlinespace
\multicolumn{4}{@{}l}{\textbf{ArBanking77 (Palestinian)}} \\
Palestinian & Jev 1.13.0 & +0.55 & [-1.06, +2.17] \\
Palestinian & Gemma-4-31B & -3.55 & [-5.60, -1.50] \\
Palestinian & Jais-2-8B & -4.13 & [-6.34, -1.92] \\
\addlinespace
\end{longtable}

\endgroup

\subsection{Identical-Input Sensitivity}\label{app:rq2-nonidentical}
The exploratory sensitivity excludes a source--variety pair when the sets of complete model-independent state/question hashes are identical on its MSA and dialect sides. It does not exclude a pair merely for sharing a passage or some answer options. No pairs are removed from Belebele or Palestinian ArBanking77; SIB-200 and DialectalArabicMMLU contain some identical inputs. The same mask applies to all four systems. \Cref{tab:rq2-nonidentical} shows Jev~1.14.0, and the companion supplies every system's restricted summaries. Seeds 13900--13903 and \RQTwoExtensionDraws{} whole-cluster draws produce separate pointwise intervals. No simultaneous or equivalence claim is based on this sensitivity.

\begin{table}[!htbp]
\centering\small
\begin{tabularx}{\linewidth}{@{}Xrrrr@{}}
\toprule
Dataset & Removed & Retained & Full gap & Nonidentical gap [95\% interval] \\
\midrule
Belebele & 0 & 4{,}500 & -5.69 & -5.69 [-6.78, -4.58] \\
SIB-200 & 795 & 7{,}237 & -0.35 & -0.39 [-1.14, +0.40] \\
DialectalArabicMMLU & 120 & 15{,}555 & -3.63 & -3.66 [-4.35, -2.96] \\
ArBanking77 (Palestinian) & 0 & 3{,}073 & -7.94 & -7.94 [-9.26, -6.59] \\
\bottomrule
\end{tabularx}

\caption{Jev~1.14.0 after removal of identical full-input pairs. Gaps and pointwise 95\% intervals are in percentage points; remaining pairs still retain source/article clustering. This is a selected nonidentical-input population, not an estimate of performance on all dialect text.}
\label{tab:rq2-nonidentical}
\end{table}

\section{Reporting Corrections and Reproducibility}
\subsection{Incomplete Primary Heads}\label{app:primary-completeness-correction}
Two OpenAI Decisions AraTox responses explicitly refused one required binary head each. The original pivot counted the incomplete items and converted missing gold to true. We reproduced that arithmetic, then applied the complete-primary-item rule used for the main reporting: 298 items remain. \Cref{tab:primary-completeness-correction} shows the effect on score and coverage. Exactly one of 2,881 system--task cells changes; all other cells and original hypothesis-test outputs are preserved. The increase reflects a changed denominator, not model improvement.
\begin{table}[htbp]
\centering\small
\begin{tabularx}{\linewidth}{@{}Xrrrr@{}}
\toprule
Reporting & AraTox items & Cov. (\%) & AraTox mean & Overall mean \\
\midrule
Original saved export & 300 & 100.00 & 61.65 & 68.78 \\
Complete primary items & 298 & 99.33 & 62.00 & 68.79 \\
\bottomrule
\end{tabularx}

\caption{Original and corrected OpenAI Decisions reporting. Scores are multiplied by 100. Original coverage is the saved export's value; the corrected row requires all primary heads.}
\label{tab:primary-completeness-correction}
\end{table}

\subsection{Source-Aware Dialect Intervals}\label{app:dialect-interval-correction}
The four matched populations were reconstructed from pinned sources and existing Jev~1.14.0 responses. Request keys, required answers, per-variety means and pair counts match the saved summaries. Belebele and SIB-200 sources map to shared FLORES articles; ArBanking77 averages available paraphrases within source and variety before pairing.

The original Belebele/SIB-200 intervals grouped by question-source ID rather than article. The first descriptive correction resampled whole articles for those datasets and whole source identifiers for the other two, using 2,000 draws initialized with seed 0. \Cref{tab:dialect-interval-correction} preserves those historical endpoints and their correction; point estimates were unchanged. The expanded panel in the main text now uses the separate \RQTwoExtensionDraws{}-draw procedure in \cref{app:rq2-extension-method}, so its Monte Carlo endpoints differ slightly. Neither procedure repairs the separate historical H2d comparison.
\begin{table}[htbp]
\centering\small
\begin{tabularx}{\linewidth}{@{}Xrr@{}}
\toprule
Dataset & Original saved interval & Recomputed interval \\
\midrule
Belebele & [-6.73, -4.56] & [-6.82, -4.63] \\
SIB-200 & [-1.02, +0.35] & [-1.02, +0.36] \\
DialectalArabicMMLU & [-4.29, -2.93] & [-4.35, -2.93] \\
ArBanking77 (Palestinian) & [-9.27, -6.61] & [-9.27, -6.61] \\
\bottomrule
\end{tabularx}

\caption{Historical original and first corrected Jev~1.14.0 interval endpoints, in percentage points. The expanded main-text panel uses more draws and different fixed seeds; matched populations and point estimates agree.}
\label{tab:dialect-interval-correction}
\end{table}

\subsection{Numerical Validation}\label{app:full-primary-validation}
All 43 frozen task counts and primary gold domains were reconstructed from recorded raw files and pinned cached datasets. Exact model/state/question keys linked inputs to existing responses, with duplicate-response and required-head checks. Recomputed primary scores, counts and coverage agree for all 2,881 cells: 2,842 finite values and 39 cells with no observed primary items. The maximum finite difference was at numerical rounding precision. This made no model calls or changes to the original databases. The companion validation report preserves the checks and hashes; intermediate cache-recovery logs are kept outside the paper. The audit does not certify source annotations, historical backend settings or original hypothesis-test inference.

\subsection{Scope and Original Study Records}\label{app:paper-scope}\label{app:hypothesis-scope}\label{app:extended-rq2}
The two paper questions are descriptive and were selected after evaluation from the broader preregistered study. The added per-dialect model panel and its uncertainty procedure were also specified after the original evaluation. This scope does not change the original hypotheses, populations or correction families. The companion \texttt{legacy\_rq1\_rq2.json} retains the original test outputs, including null and contrary outcomes, and \texttt{legacy\_analysis\_status.md} states their qualifications. Original RQ1 resampling, H1d contrast/correction rules and some original RQ2 inferential checks remain unresolved. The historical H2d effect remains withheld because its original join omitted task identity. The new task-aware descriptive panel uses independently reconstructed pairs and a separately stated comparator/uncertainty rule; it does not validate or replace that historical effect or its correction family. The deposited preregistration and original outputs are unchanged.

\end{document}